\documentclass[conference]{IEEEtran}
\IEEEoverridecommandlockouts

\usepackage{cite}
\usepackage{amsmath,amssymb,amsfonts}
\usepackage{algorithmic}
\usepackage{graphicx}
\usepackage{textcomp}
\usepackage{xcolor}
\def\BibTeX{{\rm B\kern-.05em{\sc i\kern-.025em b}\kern-.08em
    T\kern-.1667em\lower.7ex\hbox{E}\kern-.125emX}}

\graphicspath{{pic/}}
\usepackage{mathrsfs}
\usepackage{amsmath, bm}
\usepackage[linesnumbered, ruled, vlined]{algorithm2e}
\SetKwRepeat{Do}{do}{while}%
\usepackage{booktabs}
\usepackage{caption}
\usepackage{subfigure} 
\usepackage{subcaption} 

\begin{document}

\title{Approximate Borderline Sampling using Granular-Ball for Classification Tasks\\
\thanks{This work was supported by XXX.}
\thanks{*: Corresponding author.}
}

\author{\IEEEauthorblockN{1\textsuperscript{st} Qin Xie}
\IEEEauthorblockA{\textit{Chongqing Key Laboratory of Computational Intelligence} \\
\textit{Chongqing University of Posts and Telecommunications}\\
Chongqing, China \\
d210201029@stu.cqupt.edu.cn}
\and
\IEEEauthorblockN{2\textsuperscript{nd} Qinghua Zhang*}
\IEEEauthorblockA{\textit{Chongqing Key Laboratory of Computational Intelligence} \\
\textit{Chongqing University of Posts and Telecommunications}\\
Chongqing, China \\
zhangqh@cqupt.edu.cn}
\and
\IEEEauthorblockN{3\textsuperscript{rd} Shuyin Xia}
\IEEEauthorblockA{\textit{Chongqing Key Laboratory of Computational Intelligence} \\
\textit{Chongqing University of Posts and Telecommunications}\\
Chongqing, China \\
xiasy@cqupt.edu.cn}
}

\maketitle

\begin{abstract}
Data sampling enhances classifier efficiency and robustness through data compression and quality improvement. Recently, the sampling method based on granular-ball (GB) has shown promising performance in generality and noisy classification tasks. However, some limitations remain, including the absence of borderline sampling strategies and issues with class boundary blurring or shrinking due to overlap between GBs. In this paper, an approximate borderline sampling method using GBs is proposed for classification tasks. First, a restricted diffusion-based GB generation (RD-GBG) method is proposed, which prevents GB overlaps by constrained expansion, preserving precise geometric representation of GBs via redefined ones. Second, based on the concept of heterogeneous nearest neighbor, a GB-based approximate borderline sampling (GBABS) method is proposed, which is the first general sampling method capable of both borderline sampling and improving the quality of class noise datasets. Additionally, since RD-GBG incorporates noise detection and GBABS focuses on borderline samples, GBABS performs outstandingly on class noise datasets without the need for an optimal purity threshold. Experimental results demonstrate that the proposed methods outperform the GB-based sampling method and several representative sampling methods. Our source code is publicly available at https://github.com/CherylTse/GBABS.
\end{abstract}


\section{Introduction}
Data sampling plays a pivotal role in supervised machine learning, particularly for classification tasks. It offers a multitude of benefits, including reduced computational complexity, balanced class distributions, diminished effects of noise and outliers, alleviation of overfitting, and enhanced model interpretability. Over the past few decades, sampling has achieved significant advancements for classification tasks, which can be summarized into three categories: sampling methods for specific classifiers, sampling methods for specific datasets, and general sampling methods.

Sampling methods for specific classifiers leverage various aspects of the classifier, including model parameters and classification results, to guide the sampling process, allowing the classifier to actively inform sample selection to potentially improve performance. For instance,  a sampling method \cite{17} is proposed to enhance the robustness of streaming algorithms against adversarial attacks. Zhang et al.\cite{18} employ feedback from each weak classifier in ensemble learning to sample based on loss or probability scores. While these methods offer advantages in targeted training, a potential drawback lies in their inherent coupling to specific classifiers, which limits their generalizability.

Sampling methods for specific datasets aim to tailor the sampling process to the unique characteristic of a dataset, potentially improving the quality of the training dataset to improve the model's performance, including modal-specific datasets and imbalanced datasets. First, sampling methods tailored for modal-specific datasets encompass a variety of techniques. These include methods designed for text data\cite{19}, image data\cite{18,24}, point cloud data\cite{20,21}, audio data\cite{22}, and time series data\cite{23}. Second, sampling methods addressing imbalanced datasets with skewed class distributions aim to rectify the imbalance between different classes. These methods\cite{45,46} help mitigate issues such as overfitting on the majority class and underfitting on the minority class\cite{28}. Commonly employed methods in this domain include the Synthetic Minority Over-sampling Technique (SMOTE) and its variants\cite{5,27,29,40}, as well as Tomek Links\cite{26}.
However, these methods are often coupled with specific datasets. Moreover, oversampling methods such as SMOTE may blur class boundaries and increase redundancy in the sampled dataset, while undersampling methods like Tomek links may discard critical samples necessary for the classifier.

General sampling methods are those that are applicable to various types of datasets and classifiers, including simple random sampling (SRS)\cite{13}, systematic random sampling\cite{14}, stratified sampling\cite{15}, and Bootstrapping\cite{9}. These methods offer broad applicability across various machine learning tasks. However, they typically perform sampling based on the overall probability distribution, making them more susceptible to noise than other sampling methods.
As a new paradigm for processing diverse large-scale datasets, granular computing (GrC)\cite{30} can significantly improve computing efficiency by transforming complex datasets into information granules, which serve as the computing units instead of individual samples. Granular-ball computing (GBC)\cite{1} is a new branch of GrC that uses the granular-ball (GB) to represent the information granule. Inspired by GBC, Xia et al.\cite{3} propose the GB-based sampling (GBS) method that can be used for various datasets and classifiers and performs well in class noise classification. GBS addresses the limitations of the aforementioned sampling methods. Although GBS performs well, it still suffers from several limitations, as follows.
1) Existing definition of the GB cannot fully describe the positional information of all samples it contains.
2) Existing granular-ball generation (GBG) methods suffer from the issue of overlap between GBs.
3) Existing GBG methods are sensitive to purity thresholds to achieve robustness, and selecting the optimal threshold is time-consuming.

Notably, effective classification hinges on learning accurate class boundaries, such as separation points, lines, curves, surfaces, or hypersurfaces, depending on the dimensionality of the data. Borderline samples residing on these boundaries hold particular significance for training classifiers. There have been some borderline sampling methods\cite{31,34,35}. Still, they suffer from limitations: classifier-specific and computationally expensive (at least quadratic time complexity) due to their reliance on original samples as the computing unit.

As discussed, although much effort has been dedicated to sampling for classification tasks, a general and efficient sampling method for borderline samples is still lacking.
To address the aforementioned limitations, inspired by the GrC, this paper proposes an approximate borderline sampling method using GBs for classification tasks, including the restricted diffusion-based granular-ball generation (RD-GBG) method and GB-based approximate borderline sampling (GBABS) method. The main contributions are as follows.

\begin{itemize}
\item [1)] The proposed RD-GBG method eliminates GB overlap and redefines GBs, ensuring that the distribution of generated GBs aligns more closely with the original dataset.
\item [2)] The proposed RD-GBG method incorporates noise detection without searching for an optimal threshold to achieve adequate noise tolerance, thereby enhancing sampling efficiency and quality.
\item [3)] The proposed GBABS method adaptively identifies borderline samples,  reducing both class noise and redundancy in the sampled dataset, while its linear time complexity accelerates classifiers.
\end{itemize}

The remainder of this paper is organized as follows. Section \ref{notation} gives some commonly used notations. Section \ref{sec: RW} reviews related works on GBC and GBS. In Section \ref{sec: approach}, the RD-GBG and GBABS are introduced in detail. The performance of the proposed methods is demonstrated in Section \ref{sec: experiments}. Finally, the conclusion and further work are presented in Section \ref{sec: conclusion}.

\section{Notations}
\label{notation}
To make the paper more concise, in the subsequent content, let $D(D=\{(\bm{x}_1,y_1),(\bm{x}_2,y_2),\cdots,(\bm{x}_N,y_N)\})$ be a dataset, where $\bm{x}_i \in \chi \subseteq \mathcal{R}^p $ is the feature vector of the sample $(\bm{x}_i,y_i)$, $y_i(y_i\in \mathcal{Y}, \mathcal{Y}=\{l_1,l_2,\cdots,l_q\})$ is its class, and $i=1,2,\cdots,N$.
The low-density sample set is denoted as $L (L \subseteq D)$.
Samples that have not been divided into any GB are called undivided samples. The set of undivided samples is denoted as $U$.
$G$ is the set of GBs generated on $D$, where $G=\{gb_1,gb_2,\cdots,gb_m\}$, $\mathbb{C}=\{(\bm{c}_1,l_1),(\bm{c}_2,l_2),\cdots,(\bm{c}_m,l_m)\}$ is the corresponding center set. The sampled dataset is denoted as $S$. Furthermore, samples of the same class are called homogeneous samples; otherwise, they are called heterogeneous samples, and the same applies to GB.

\section{Related Work}
\label{sec: RW}

\subsection{Granular-Ball Computing}
GBC\cite{1} is a family of scalable, efficient, and robust data mining methods, which is a two-stage learning, including the GBG stage and the GB-based learning stage. The core idea of the GBC is to employ the ball of varying granularity to represent the information granule and replace the sample for calculating in various tasks. The geometry of the ball is completely symmetrical, and only the center and radius are required to characterize it in any dimension, so it can be easily applied to diverse scenarios, including classification\cite{1,2,48}, clustering\cite{36,37,47}, fuzzy sets\cite{39,41}, feature engineering\cite{42,43,44} and deep learning\cite{38}.

As a granulation method, the core idea of the GBG method is to cover a dataset with a set of balls, where a ball is called a GB $gb=(O,(\bm{c},r,\mathbb P,l))$. Specifically, the granulation process of the existing GBG methods can be briefly described as follows. First, the whole training dataset is initialized as the initial GB. Second, $k$-means\cite{1}, $k$-division\cite{2}, or hard-attention division\cite{33} is employed to split the GB into $k$ or more finer GBs. The center $\bm{c}_i$ and radius $r_i$ of $gb_i(\forall gb_i \in G, i=1,2,\cdots,m)$ are defined as follows.
\begin{equation}\label{eq1}
\bm{c}_i=\frac{1}{|D_i|}\sum_{(\bm{x},y)\in D_i}\bm{x},
\qquad
r_i=\frac{1}{|D_i|}\sum_{(\bm{x},y) \in D_i} \bigtriangleup(\bm{x},\bm{c}_i),
\end{equation}
where $D_i\in D$, $|\bullet|$ represents the cardinality of set $\bullet$, and $\bigtriangleup(\cdot,\star)$ denotes the distance function. Without losing generality, Euclidean distance is employed in this paper.
For most real datasets, the samples are unevenly distributed in the feature space, and the GB defined by Eq.\ref{eq1} will cause some samples to be distributed outside the ball.

The label $l_i$ of the $gb_i$ is determined by the majority of samples contained within it.
The quality of the $gb_i$ is measured using the purity $\mathbb P_i$, that is, the ratio of the number of samples within the $gb_i$ that are consistent with its label $l_i$ to the number of all samples within it.
The closer the purity of GB is to $1.0$, the closer the distribution of GBs is to the original dataset. Iteratively split each GB until the purity of each GB reaches the given purity threshold.

Existing GBG methods suffer from overlapping GBs, causing the distribution of GB sets to diverge from that of the original dataset, leading to inconsistency between the sampled and original datasets. Although this issue tends to alleviate with increasing purity\cite{1}, it cannot be fully resolved.
For instance, overlapping heterogeneous GBs would blur class boundaries, and overlapping homogeneous GBs can cause the shrinking of class boundaries.

\subsection{GB-based Sampling Method}

Inspired by GBC, a general GB-based sampling method (GGBS) and a GB-based sampling method for imbalanced datasets (IGBS) are proposed by Xia et al.\cite{3}, both including the GBG stage and the undersampling stage.

The core idea of the GBG method used in the GBG stage of GGBS and IGBS can be briefly described below. Given a dataset $D$, it is initialized to the initial GB. For each GB, if its purity is less than the purity threshold and the number of samples within the GB is greater than $2\times p$, then the $k$-division is used to split the GB into $k$ finer GBs. Iteratively, until the purity of each GB reaches the threshold or the number of samples it contains is less than or equal to $2\times p$. Finally, a GB set $G$ is obtained. In this section, a GB is called a small GB if it contains no more than $2\times p$ samples; otherwise, it is called a large GB.

The core idea of the undersampling stage of GGBS can be summarized as follows. First, all samples contained in small GBs are put into the sampled dataset $S$. Second, for each large GB, put $2\times p$ samples into the sampled dataset $S$, which are the homogeneous sample closest to the intersection point of the GB and the coordinate in each feature dimension.

The core steps of the undersampling stage of IGBS are as follows. First, the first step is the same as that for GGBS. Second, for each minority class GB that is large, all the containing minority class samples are sampled into $S$. Third, for each majority class GB that is large, $2\times p$ majority class samples are sampled into $S$, whose sampled rule is the same as GGBS. Finally, if the class distribution is still skewed, randomly sample more majority samples into $S$.

However, the aforementioned GBG method stops splitting GBs to ensure a preset sample count, even if the purity threshold is unmet, and GB overlaps further degrade their quality. These issues reduce the quality of the sampled data in GGBS and IGBS. Additionally, GGBS applies a uniform sampling strategy across all GBs, ignoring the importance of borderline GBs, which may retain redundancy or noise, limiting classifier improvement. Moreover, IGBS blindly balances class ratios without assessing sample redundancy, increasing the risk of overfitting.

\section{Approach}
\label{sec: approach}
The proposed sampling method is a two-stage learning approach, namely, the GBG stage and the GB-based sampling stage. This section will introduce the proposed RD-GBG method and the GBABS method, respectively.

\subsection{Framework}
\label{framework}

\begin{figure*}[htbp]
\vspace{-0.3in}
\centering
\includegraphics[height=3.0in,width=7in]{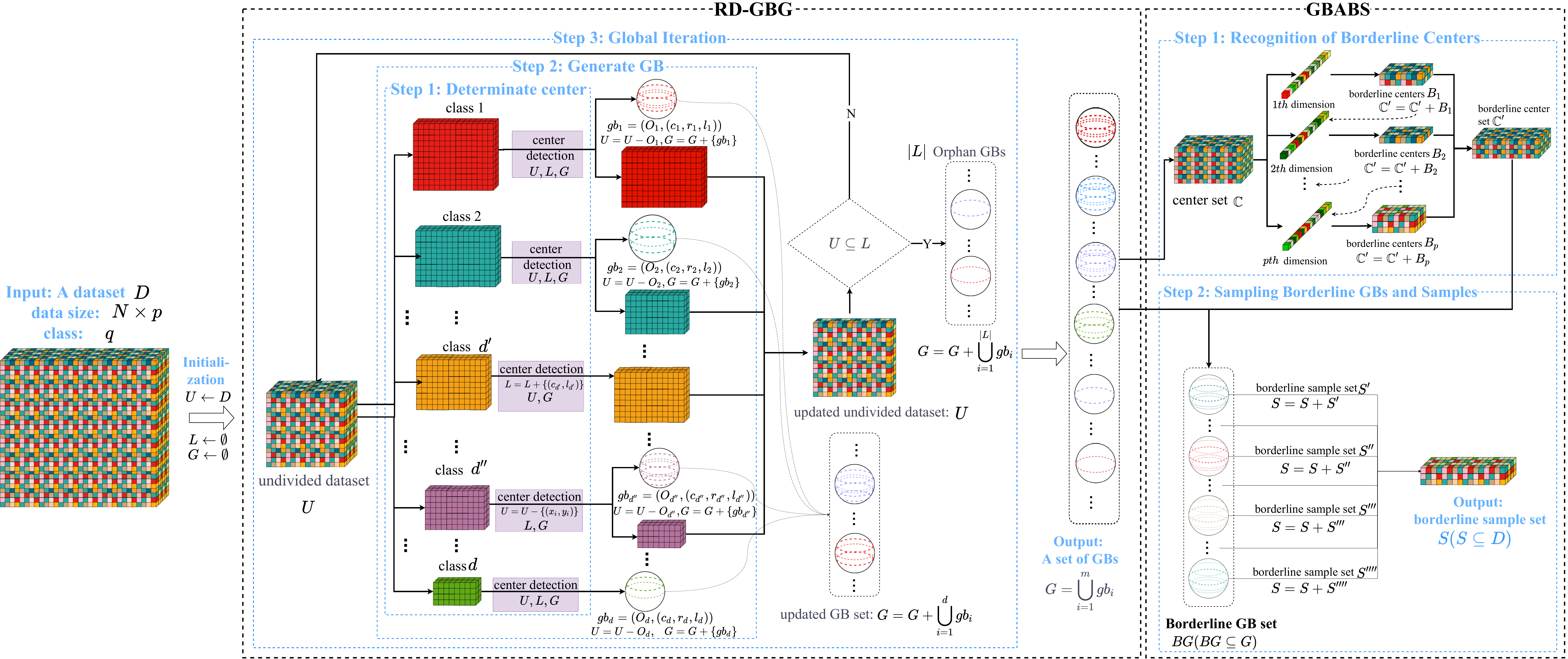}
\caption {Architecture of GBABS based on RD-GBG method.}
\label{fig1}
\vspace{-0.1in}
\end{figure*}

The architecture of the RD-GBG method is shown on the left side of Fig.\ref{fig1}. The entire training dataset is initialized as the undivided sample set. First, the undivided sample set is grouped by labels, and a sample is randomly chosen as the candidate center from each group, prioritizing larger groups. And perform center detection to determine whether the center meets the local consistency which means that the center has neighbors that are homogeneous with it. Second, construct the pure GB based on each eligible center on the undivided sample set, as well as the new GB cannot overlap with the previous GBs. Iteratively, the above process is performed on the undivided sample set until the undivided samples with local consistency converge. Lastly, orphan GBs are constructed.

The architecture of the GBABS method is shown on the right side of Fig.\ref{fig1}. First, the RD-GBG method is performed for a given dataset to obtain a GB set. Second, take the centers of all GBs to form a center set to represent their location information in the feature space. Third, based on the center set, the GBs on the class boundaries are detected from each feature dimension. Finally, sampling is performed based on the heterogeneous adjacent relation between borderline GBs.

\subsection{Restricted Diffusion-based GBG Method}
\label{RDA-GBG}
In the field of GrC, the granulation method for large-scale datasets needs to follow three criteria. The first one is that the distribution of information granules should be as consistent as possible with that of the original dataset, which can be called approximation. The second one is that a GB should contain as many samples as possible to improve the efficiency as well as ensure the performance of the GB-based downstream learning tasks, which can be called representativeness. The third one is that the samples should be used as much as possible, which can be called completeness.

Consequently, based on the idea of restricted diffusion, a new GBG method is proposed in this section, which is adaptive and without overlap among GBs.
As shown in Fig.\ref{fig1}, the whole training dataset $D$ is initialized to the undivided sample set $U$. The GB is constructed on $U$ in turn iteratively. Specifically, the construction process of GB will be introduced in detail below, which includes the determination method of local-density centers, the construction method of the GB, and the iteration termination condition.

\subsubsection{Determination Method of Local-density Centers}
\label{DMLDC}
Considering that the center of the GB should be representative and the method should apply to datasets of different shapes, the center is selected randomly with local consistency; namely, at least the nearest neighbor is homogeneous to the center. A method for determining the local-density center is proposed below, as Step 1 of RD-GBG module of Fig.\ref{fig1}.

Suppose the potential center set $U-L$ denoted as $T$, where $T=\{T_1,T_2,\cdots,T_d\}$, $\bigcap T_i = \emptyset$, $\bigcup T_i = T$, $|T_1| \geq |T_2| \geq \cdots \geq |T_d|$, $d(d \le q)$ represents the number of class in $T$, all samples in $T_i$ are homogeneous, and $i=1,2,\cdots,d$.

Randomly select a sample denoted as $(\bm{c_i},l_i)$ from each $T_i$ to form a candidate center sequence $\bm{C}_{cand}=\{(\bm{c}_1,l_1),(\bm{c}_2,l_2),\cdots,(\bm{c}_d,l_d)\}$.
Since each element in $\bm{C}_{cand}$ is randomly selected, they may not satisfy the local consistency. Therefore, the elements in $\bm{C}_{cand}$ need to be detected to obtain eligible centers, called local-density centers. A detailed introduction to the detection method is below.

For $\forall (\bm{c},l) \in \bm{C}_{cand}$, calculate the distance $\bigtriangleup(\bm{x}_i,\bm{c})$ between $(\bm{c},l)$ and each $(\bm{x}_i,y_i) \in U-\{(\bm{c},l)\}$, $i=1,2,\cdots,|U|-1$.
If the sample $(\bm{x},y)$ closest to $(\bm{c},l)$ is homogeneous with it, then $(\bm{c},l)$ is a local-density center, otherwise further check the number $h(\bm{c},l)$ of samples that are heterogeneous with it in $\rho$ nearest neighbors $\bm{N}_\rho(\bm{c},l)$, $\bm{N}_\rho(\bm{c},l)\subseteq U$.

\begin{equation}\label{eq6}
h(\bm{c},l) = |\{(\bm{x},y)|(\bm{x},y) \in \bm{N}_\rho(\bm{c},l),y \neq l\}|,
\end{equation}
where $\rho$ refers to density tolerance.

The local-density center detection rules are as follows, where the local-density center sequence is denoted as $\bm{C}$.
\begin{itemize}
\item If $h(\bm{c},l)=\rho$, then $(\bm{c},l)$ is judged as a class noise and update $U$ to $U-\{(\bm{c},l)\}$;
\item If $h(\bm{c},l)=1$, then the nearest neighbor sample $(\bm{x},y)$ is determined as a class noise and update $U$ to $U-\{(\bm{x},y)\}$. Update $\bm{C}$ to $\bm{C}+\{(\bm{c},l)\}$;
\item If $1<h(\bm{c},l)<\rho$, namely, $(\bm{c},l)$ cannot be distinguished from other classes to be judged as a low-density sample, then update $L$ to $L+\{(\bm{c},l)\}$.
\end{itemize}
Consequently, there are the local-density center sequence $\bm{C}=\{(\bm{c}_1,l_1),(\bm{c}_2,l_2),\cdots,(\bm{c_}{d^{\prime}},l_{d^{\prime}})\}$, $d^{\prime} \leq d$, the updated undivided sample set $U$ and low-density sample set $L$.

\begin{figure}[htbp]
\centering
\includegraphics[height=2.5in,width=2.5in]{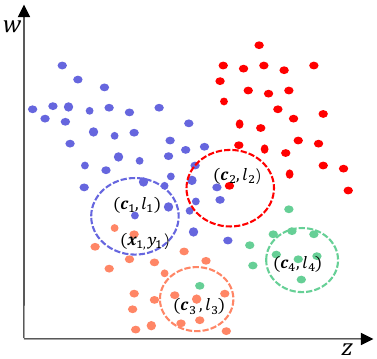}
\caption {Example for detecting local-density centers.}

\label{fig2}
\end{figure}

As shown in Fig.\ref{fig2}, there is a dataset $D$ with 4 classes marked in different colors. First, the undivided sample set $U$ is initialized to $D$, the low-density sample set $L$ is initialized to $ \emptyset$, and the potential center sample set $T$ is $U$, $T=\{T_1,T_2,T_3,T_4\}$, where $|T_1|=42,|T_2|=26,|T_3|=20,$ and $|T_4|=10$. Second, $(\bm{c}_1,l_1),(\bm{c}_2,l_2),(\bm{c}_3,l_3),$ and $(\bm{c}_4,l_4)$ are randomly selected heterogeneous centers to form a candidate center sequence $\bm{C}_{cand}=\{(\bm{c}_1,l_1),(\bm{c}_2,l_2),(\bm{c}_3,l_3),(\bm{c}_4,l_4)\}$. Third, to confirm whether these centers satisfy local consistency, local-density center detection is performed on each $(\bm{c}_i,l_i) \in \bm{C}_{cand}$ respectively. Without losing generality, let the density tolerance $\rho$ be $5$. For $(\bm{c}_1,l_1)$, since its nearest neighbor $(\bm{x}_1,y_1)$ is heterogeneous with it and some other its 5 nearest neighbors are homogeneous with it, $(\bm{c}_1,l_1)$ can be considered as a low-density sample rather than a qualified center. For $(\bm{c}_2,l_2)$, since its 5 nearest neighbors are heterogeneous with it, then $(\bm{c}_2,l_2)$ is a class noise. Moreover, for $(\bm{c}_3,l_3)$, since its nearest sample is heterogeneous with it and the others in 5 nearest neighbors are all homogeneous with it, $(\bm{c}_3,l_3)$ can be taken as the eligible center, and its nearest sample is identified as a class noise. In addition, for $(\bm{c}_4,l_4)$, since it is homogeneous with its 5 nearest neighbors, it can be taken as an eligible center. As a result, the local-density center sequence is $\bm{C}=\{(\bm{c}_3,l_3), (\bm{c}_4,l_4)\}$.

\subsubsection{Generation Method of the GB}
\label{GBG}
When other conditions remain unchanged, the greater the purity of the GB, the more consistent the distribution of GBs is with the distribution of the original dataset. Thus, all the purity of generated GBs is 1.0, namely, pure GBs. To construct pure GB with more samples without overlap, consider the centers in local-density center sequence $\bm{C}$ sequentially and adopt a strategy of diffusion from the center and stopping when encountering heterogeneous samples or previous generated GBs.
A method for generating the GB without overlapping is proposed below, as Step 2 of RD-GBG module of Fig.\ref{fig1}.

First, suppose that a set of GBs $G^\prime =\{gb_1,gb_2,\cdots,gb_{m^{\prime}}\}$ has been generated on $D-U$, $m^{\prime} \le m$.
For each $(\bm{c},l) \in \bm{C}$, calculate the distance between $(\bm{c},l)$ and each $(\bm{x}_i,y_i) \in U-\{(\bm{c},l)\}$. If the $(\omega +1)th$ nearest neighbor of $(\bm{c},l)$ is heterogeneous with it and the $\omega$ nearest neighbors are all homogeneous with it, the distance corresponding to the $\omega th$ nearest neighbors is called locally consistent radius of $(\bm{c},l)$, denoted as $CR(\bm{c})$.

\begin{equation}\label{eq11}
CR(\bm{c}) = \max \{\bigtriangleup(\bm{c}^\prime,\bm{c})|(\bm{c}^\prime,l) \in \bm{N}_\omega (\bm{c},l)\},
\end{equation}
where the label of any sample in $\bm{N}_\omega (\bm{c},l)$ is $l$, and there is a sample in $\bm{N}_{\omega+1}(\bm{c},l)$ whose label is not $l$.

Second, to ensure there is no overlap between GBs, the distance from the center $(\bm{c},l)$ to the nearest constructed GB should be considered, denoted as the conflict radius $r_{conf}(\bm{c})$.
\begin{equation}\label{eq7}
r_{conf}(\bm{c}) = \min_{i=1,2,\cdots,m^{\prime}} \{\bigtriangleup(\bm{c}_i,\bm{c})-r_i\},
\end{equation}
where $\bm{c}_i$ and $r_i$ are center and radius of $ gb_i \in G^\prime$, respectively.

Notably, if $r_{conf}(\bm{c}) < CR(\bm{c})$, the distance corresponding to the sample in $\bm{N}_\omega (\bm{c},l)$ farthest from $(\bm{c},l)$ without overlapping with previous GBs should be taken as the radius, which called the restricted maximum consistent radius $r_{max}(\bm{c})$.
To summarize, the radius $r$ can be represented as follows.
\begin{equation}\label{eq9}
r = \begin{cases}
& CR(\bm{c}),\text{ if } CR(\bm{c})\le r_{conf}(\bm{c}), \\
& r_{max}(\bm{c}), \text{ if } CR(\bm{c}) >  r_{conf}(\bm{c}),
\end{cases}
\end{equation}
where $r_{max}(\bm{c})$ is defined as below.
\begin{equation}\label{eq8}
r_{max}(\bm{c})=\max_{(\bm{x}_i,y_i)\in U}\{\bigtriangleup(\bm{x}_i,\bm{c})| \bigtriangleup(\bm{x}_i,\bm{c})\le r_{conf}(\bm{c})\}.
\end{equation}
If $r=0$, then the center is distributed on the edge of the undivided sample set, and the center might be divided into other GB containing multiple samples later. Therefore, only consider the case that $r\neq 0$. The set $O$ of samples that fall within a ball with $(\bm{c},l)$ as center and $r(r\neq 0)$ as the radius is defined below.
\begin{equation}\label{eq10}
O = \{(\bm{x},y)|\bigtriangleup(\bm{x},\bm{c})\leq r,(\bm{x},y)\in U-\{(\bm{c},l)\}\}.
\end{equation}
Consequently, the GB $gb=(O,(\bm{c},r,l))$ is generated. Update $G$ to $G+\{gb\}$, and $U$ to $U-O$. Iteratively, until all centers in $\bm{C}$ are considered.

Notably, all the samples in $O$ are covered by $gb$, the $gb$ can correctly represent the positional information of all samples in $O$. Long story short, the defined $\bm{c}$ characterizes the central tendency of samples in $O$, while the defined $r$ delineates the potential maximum boundary of $O$ in the feature space. This is extremely valuable for sampling tasks.

\begin{algorithm}[htbp]
	\caption{RD-GBG Method.}
	\label{alg1}
	\SetKwInOut{Input}{Input}\SetKwInOut{Output}{Output}
	\Input {Dataset $D$, Density tolerance $\rho$.}
	\Output {A set of GBs $G$.}
	$U$ represents the undivided sample set; $L$ represents the low-density sample set\;
	Initialize $G \leftarrow \emptyset$, $U \leftarrow D$, $L \leftarrow \emptyset$\;
	\Repeat{$U \subseteq L$}{
		Randomly select $d(d \le q)$ heterogeneous samples from $T(T=U-L)$ to form $\bm{C}_{cand}$\;
		\For{$(\bm{c},l)$ in $\bm{C}_{cand}$}{
			Calculate the distances between $\bm{c}$ and each sample in $U$\;
            Obtain the nearest neighbor $(\bm{x},y)$ of $(\bm{c},l)$\;
			\If {$y \neq l$}{
                Get the $h(\bm{c},l)$ by Eq.\ref{eq6}\;
				\If{$h(\bm{c},l)==\rho$}{
				$U \leftarrow U-\{(\bm{c},l)\}$\;
				\textbf{continue}\;
			}
				\ElseIf{$h(\bm{c},l)==1$}{
				$U \leftarrow U - \{(\bm{x},y)\}$\;
			}
				\Else{
					$L \leftarrow L + \{(\bm{c},l)\}$\;
					\textbf{continue}\;
			}
        }
			Obtain locally consistent radius $CR(\bm{c})$ Eq.\ref{eq11}\;
			Obtain conflict radius $r_{conf}(\bm{c})$ with $G$ by Eq.\ref{eq7}\;
			\If{$CR(\bm{c}) <= r_{conf}(\bm{c})$}{
            $r \leftarrow CR(\bm{c})$\;
		}
		      \Else{
                Calculate $r_{max}(\bm{c})$ by Eq.\ref{eq8}\;
                $r \leftarrow r_{max}(\bm{c})$\;
		}
            \If{$r\neq 0$}{
                Obtain sample set $O$ by Eq.\ref{eq10}\;
                $gb=(O,(\bm{c},r,l))$\;
			     $U \leftarrow U - O$;
			     $G\leftarrow G + \{gb\}$\;
            }
            \Else{$L \leftarrow L + \{(\bm{c},l)\}$\;}
			}
	}
        Generate the orphan GB on $U$ to obtain GB set $OG$\;
	$G \leftarrow G + OG$\;		
	Return $G$.
\end{algorithm}

\begin{figure}[htbp]
\centering
\includegraphics[height=2.5in,width=2.5in]{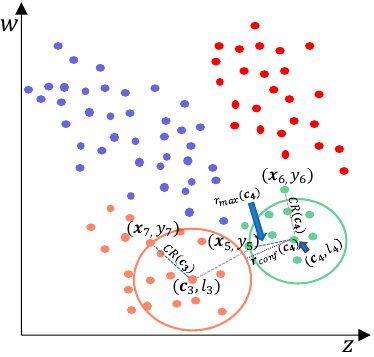}
\caption {Example for generation of GB.}
\label{fig3}
\end{figure}

As shown in Fig.\ref{fig3}, based on the Section \ref{DMLDC}, the local density-center sequence $\bm{C} = \{(\bm{c_3},l_3),(\bm{c_4},l_4)\}$ is obtained based Fig.\ref{fig2}. Due to $|T_3|>|T_4|$, it is preferred to construct GB centered $(\bm{c}_3,l_3)$ on $U$. Based on the distances between $\bm{c}_3$ and $\bm{x}_i((\bm{x}_i,y_i) \in U-\{(\bm{c}_3,l_3)\}$, it can be found that the 11 nearest neighbors of $(\bm{c}_3,l_3)$ is homogeneous with it, while the $12th$ nearest neighbor is not. Therefore, the locally consistent radius $CR(\bm{c}_3)$ is the distance between $\bm{c}_3$ and $\bm{x}_7$. There is no previous GBs, then $r_3 = CR(\bm{c}_3)$, $O_3 = \{(\bm{x},y)\in U \mid \bigtriangleup(\bm{c}_3,\bm{x}) \leq r_3\}$. As a result, the GB constructed on $U$ centered around $(\bm{c}_3,l_3)$ is assembled as $gb_1 = \{\bm{c}_3, r_3, l_3, O_3\}$.
Similarly, construct a new GB centered $(\bm{c}_4,l_4)$ on $U-O_3$. The locally consistent radius $CR(\bm{c}_4)$ is the distance between $\bm{c}_4$ and $\bm{x}_6$. In addition, there is a previous GB $gb_1$, calculate the distance between $\bm{c}_4$ and $gb_1$ to obtain $r_{conf}(\bm{c}_4)$ by Eq.\ref{eq7}. Due to $r_{conf}(\bm{c}_4) \le CR(\bm{c}_4)$ and the restricted maximum consistent radius $r_{max}(\bm{c}_4)= \bigtriangleup(\bm{x}_5,\bm{c}_4) \ne 0$, then $r_4=r_{max}(\bm{c}_4)$, and the new GB is $gb_2 = (\bm{c}_4, r_4, l_4, O_4)$, where $O_4 = \{(\bm{x},y)\in U-O_3 \mid \bigtriangleup(\bm{c}_4,\bm{x}) \leq r_4\}$.

\subsubsection{Iteration Termination Condition and Time Complexity}
\label{ITC}
As shown in Step 3 of RD-GBG module of Fig. \ref{fig1}, some new GBs are generated in Step 2, and both the undivided sample set $U$ and the low-density sample set $L$ are updated. If all undivided samples are low-density samples, that is, there is no potential center, then terminate iteration. The iteration termination condition is to judge whether $U\subseteq L$ is reached.
Moreover, considering the completeness of the abovementioned granularity criteria, all low-density and undivided samples are respectively constructed as GBs with a radius of $0$. Algorithm \ref{alg1} provides the complete RD-GBG method. Notably, to avoid redundant calculations, the distance calculated by the local-density center detection method in Section \ref{DMLDC} is also used for subsequent construction of the GB in Section \ref{GBG}.

Suppose a dataset that contains $N$ samples and $q$ classes. Let $N_i(i=1,2,\cdots,t)$ represent the number of samples divided into GBs in the $ith$ iteration, and $q_i$ denotes the class number of undivided samples in the $ith$ iteration. In the $1th$ iteration, randomly select $q_1$ centers to generate GBs, and the time complexity is $O(q_1N)$. In the $2th$ iteration, randomly select $q_2$ centers to generate GBs, and the time complexity is $O(q_2(N-N_1))$. Assume that RD-GBG iterates for $t$ iterations. In the $t$th iteration, randomly select $q_t$ centers to generate GBs, and the time complexity is $O(q_t(N-N_1-\cdots-N_{t-1}))$. Notably, each iteration processes fewer undivided samples than the previous iteration. Consequently, the total time complexity is much lower than $O(tqN)$.

\subsection{GB-based Approximate Borderline Sampling}
\label{GBABS}

According to Section \ref{RDA-GBG}, the set of GBs constructed on a given dataset can essentially describe this dataset approximately, including the class boundaries of the dataset. Therefore, there are GBs distributed on the class boundaries, called borderline GBs, which can be detected somehow.
The geometric center is a geometric property of a ball that represents its position.

\begin{figure}[htbp]
\vspace{-0.3cm}
	\centering
	\begin{minipage}[t]{0.5\linewidth}
		\centering
		\includegraphics[height=3.3cm, width=4cm]{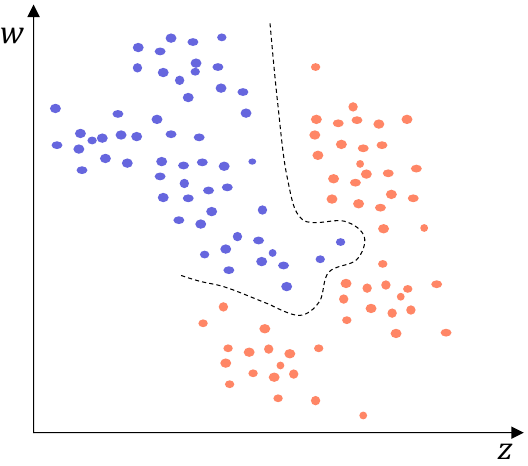}
		\small (a) Original dataset.
	\end{minipage}%
	\begin{minipage}[t]{0.5\linewidth}
		\centering
		\includegraphics[height=3.3cm, width=4cm]{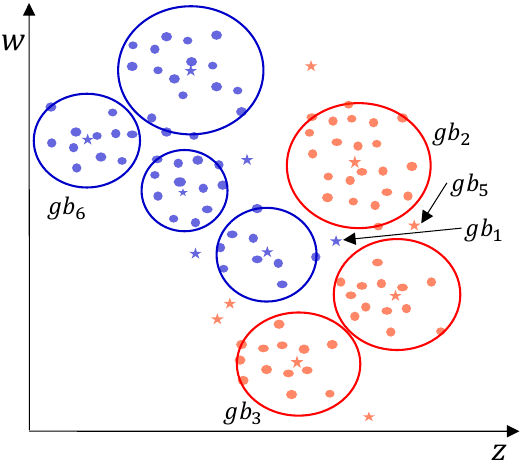}
		\small (b) All generated GBs.
	\end{minipage}%
    \quad
	\begin{minipage}[t]{0.5\linewidth}
		\centering
		\includegraphics[height=3.3cm, width=4cm]{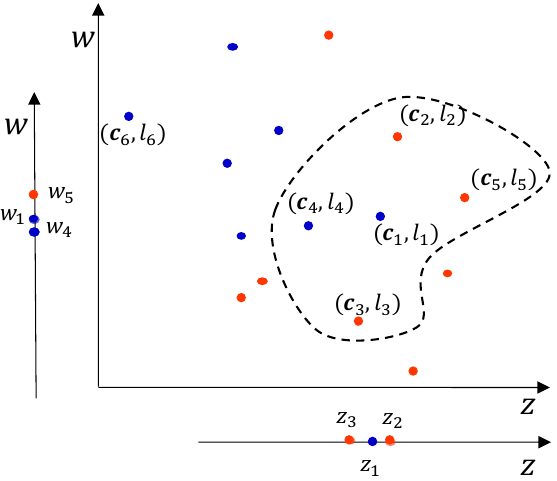}
		\small (c) Centers of all GBs.
	\end{minipage}%
	\begin{minipage}[t]{0.5\linewidth}
		\centering
		\includegraphics[height=3.3cm, width=4cm]{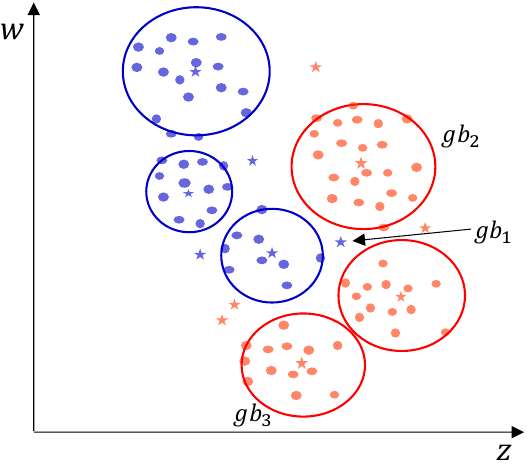}
		\small (d) Borderline GBs.
	\end{minipage}%
    \quad
    \begin{minipage}[t]{0.5\linewidth}
	\centering
	\includegraphics[height=3.3cm, width=4cm]{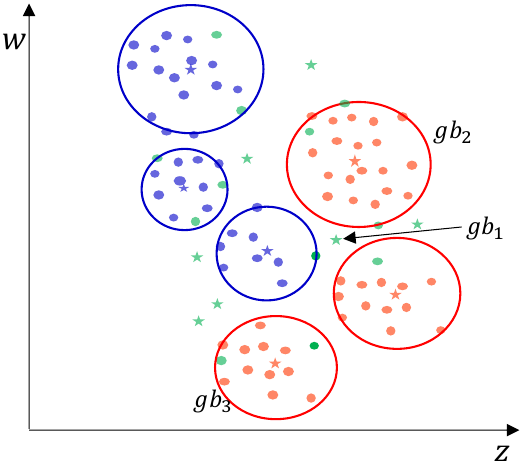}
	\small (e) Borderline GBs and samples.
    \end{minipage}%
    \begin{minipage}[t]{0.5\linewidth}
	\centering
	\includegraphics[height=3.3cm, width=4cm]{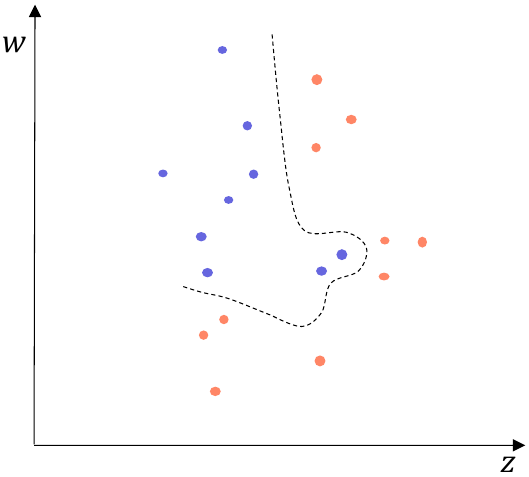}
	\small (f) Borderline samples.
    \end{minipage}%
\caption {Example for recognizing borderline GBs and samples.}
\label{fig4}
\end{figure}

Typical distance measurement methods, such as Euclidean distance, fail when determining the location of multidimensional data objects. As shown in Fig. \ref{fig4}(a), there is a two-dimensional dataset with 2 classes marked in different colors. A GB set is generated on the dataset using the RD-GBG method, shown in Fig. \ref{fig4}(b). All the centers of these GBs are shown in Fig. \ref{fig4}(c). Calculate the distance between center $(\bm{c}_1,l_1)$ and other centers, and only find that the nearest heterogeneous center is $(\bm{c}_5,l_5)$. Then it can be judged that there is only a separation point between $(\bm{c}_1,l_1)$ and $(\bm{c}_5,l_5)$. Obviously, there is also another separation point between $(\bm{c}_2,l_2)$ and $(\bm{c}_1,l_1)$. Therefore, calculating the distance between samples to identify the class boundaries will fail.

In high-dimensional feature space, the position of the sample is usually represented by multi-dimensional coordinates composed of its feature values. Therefore, the coordinates of the center of the GB are used as its positional information in the feature space.
As shown in Step 1 of the GBABS module of Fig.\ref{fig1}, for $\forall (\bm{c}_i,l_i) ((\bm{c}_i,l_i)\in \mathbb{C})$, if at least one of its left and right neighbors $(\bm{c}_j,l_j)\in \mathbb{C}$ in a given feature dimension is heterogeneous with it, it can be judged that both $(\bm{c}_i,l_i)$ and $(\bm{c}_j,l_j)$ are likely to be distributed on the class boundaries, where $i\neq j$. Consequently, the borderline center set $\mathbb{C}^{\prime}(\mathbb{C}^{\prime}\subseteq \mathbb{C})$ would be obtained when all $(\bm{c}_i,l_i)\in \mathbb{C}$ are considered.

As shown in Fig. \ref{fig4}(c), for center $(\bm{c}_1,l_1)$, in the feature column corresponding to the feature $z$, its left and left neighbors are $(\bm{c}_2,l_2)$ and $(\bm{c}_3,l_3)$, respectively. Thus, as shown in Fig. \ref{fig4}(d), the $gb_1,gb_2$ and $gb_3$ are recognized as borderline GBs, owing to that $l_1\neq l_2$ and $l_1\neq l_3$. And, in the feature column corresponding to the feature $w$, the left and left neighbors of $(\bm{c}_1,l_1)$ are $(\bm{c}_4,l_4)$ and $(\bm{c}_5,l_5)$. Since $(\bm{c}_4,l_4)$ is homogeneous with $(\bm{c}_1,l_1)$, and $(\bm{c}_5,l_5)$ is heterogeneous with it. Thus, the $(\bm{c}_5,l_5)$ is recognized as another borderline center. Notably, as shown in Fig. \ref{fig4}(c), the left and right neighbors of $(\bm{c}_6,l_6)$ in all feature dimensions are homogeneous with it, then the $(\bm{c}_6,l_6)$ can be judged as an intra-class center.

The GBs corresponding to the borderline centers are the borderline GBs. As shown in Step 2 of the GBABS module of Fig.\ref{fig1}, the borderline GB set $BG(BG\subseteq G)$ can be obtained based on borderline center set $\mathbb{C}^{\prime}$. For $\forall gb \in BG$, there is at least one sample closest to the class boundary in a certain dimension, which is a dimension that the $gb$ is judged to be a borderline GB. Consequently, the borderline sample set $S(S\subseteq D)$ can be obtained, in which there are no repeated samples.

All the borderline GBs are shown in Fig.\ref{fig4}(d). In Fig.\ref{fig4}(e), for the feature column corresponding to feature $z$, the left and right neighbors of the borderline GB $gb_1$ are $gb_2$ and $gb_3$, respectively. Therefore, the sample with the largest value of feature $z$ among the samples in $gb_3$ is identified as a borderline sample. Similarly, all green-marked samples in Fig. \ref{fig4}(e) are the borderline samples. Consequently, the sampled dataset is shown in Fig. \ref{fig4}(f), representing the approximate borderline sample set. Compared to Fig. \ref{fig4}(a), Fig. \ref{fig4}(f) exhibits a significantly reduced number of samples while maintaining essentially the same class boundaries.

Algorithm \ref{alg2} provides the complete GBABS method. Suppose a dataset $D$ that contains $N$ samples and $q$ classes with $p$ features. To obtain a GB set $G=\{gb_1,gb_2,\cdots,gb_m\}$ with Algorithm \ref{alg1}, the time complexity is $O(tqN)$. The time complexity for sampling on the $G$ using GBABS is $O(pm\log m)$. As a result, the total time complexity is $O(tqN+pm\log m)$, which is still linear.

\begin{algorithm}[htbp]
\caption{GBABS Method.}
\label{alg2}
\SetKwInOut{Input}{Input}\SetKwInOut{Output}{Output}
\Input {Dataset $D$ with $p$ features.}
\Output {Sampled dataset $S$.}
Initialize $S \leftarrow \emptyset$\;
    Generate a GB set $G=\{gb_1,gb_2,\cdots,gb_m\}$ on $D$ by Algorithm \ref{alg1}\;
    Obtain center set $\mathbb{C}=\{(\bm{c}_1,l_1),(\bm{c}_2,l_2),\cdots,(\bm{c}_m,l_m)\}$ of $G$\;
    \For{$i$ from $0$ to $p-1$}{
        Obtain all adjacent and heterogeneous GBs $gb_j,gb_k$ based on $(\bm{c}_j,l_i),(\bm{c}_k,l_k)\in \mathbb{C}$ along the $ith$ feature\;
        Obtain the adjacent samples $\bigcup \{(\bm{x},y)\}$ along the $ith$ feature in $gb_j$ and $gb_k$\;
        $S\leftarrow S+\bigcup \{(\bm{x},y)\}$\;
    }
Return $S$.
\end{algorithm}

\section{Experimental Results and Analysis}
\label{sec: experiments}
In this section, the proposed methods will be validated in terms of the effectiveness of the sampling ratio, the robustness against class noise, the effectiveness of handling imbalanced datasets, and the parameter sensitivity analysis. All experiments are conducted on a system with a 3.00GHz Intel i9-10980XE CPU and Python 3.9.7.

\subsection{Experimental Settings}

\subsubsection{Baselines}
The proposed GBABS is compared with the GGBS\cite{2}, IGBS\cite{2}, SRS\cite{13}, SMOTE (SM)\cite{4}, borderline SMOTE (BSM)\cite{5}, SMOTENC (SMNC)\cite{4}, and TomekLinks (Tomek)\cite{26} on several widely used machine learning classifiers, that is, $k$-nearest neighbor ($k$NN)\cite{12}, decision tree (DT)\cite{11}, Random Forest (RF)\cite{9}, light gradient boosting machine (LightGBM)\cite{10}, and Extreme Gradient Boosting (XGBoost)\cite{8}. Notably, the GGBS and IGBS are state-of-the-art GB-based sampling methods, whereas IGBS is specially designed for imbalanced datasets. The SM, BSM, and SMNC are representative oversampling methods for imbalanced datasets, and Tomek is the corresponding common undersampling method. The SRS is the representative unbiased general sampling method.

\subsubsection{Datasets}
Comparative experiments are conducted on diverse datasets from various domains, including finance, medical diagnosis, and handwritten digit recognition. These datasets, randomly selected from the UCI Machine Learning Repository \cite{6}, KEEL-dataset repository \cite{7}, and Kaggle, span various sample sizes, feature dimensions, and class distributions. The datasets vary from small-scale (e.g., Credit Approval) to large-scale (e.g., shuttle), low-dimensional (e.g., banana) to high-dimensional (e.g., USPS), and binary (e.g., Diabetes) to multi-class (e.g., USPS). Detailed dataset information, including imbalance ratio (IR), is provided in Table \ref{table1}, where IR represents the ratio of majority to minority class samples. Class noise datasets with noise ratios of $5\%$, $10\%$, $20\%$, $30\%$, and $40\%$ are constructed on all datasets by randomly selecting samples and altering their labels.

\subsubsection{Metrics and Parameter Settings}
The commonly used evaluation metric $Accuray$ in supervised learning is employed. The metric $G-mean$ is taken to validate the performance of the imbalanced classification. Moreover, the five-fold cross-validation method is employed to reduce the risk of overfitting, which is repeated five times to calculate the average metric value as the final result to avoid possible bias.
The scikit-learn is employed for all used classifiers, which is a popular open-source machine-learning library for Python. The parameters for all the classifiers are consistent with the default parameters in scikit-learn. Moreover, the random seeds are set in all used classifiers for a fair comparison. Notably, the sampling ratio of the SRS on each dataset is consistent with that of GBABS.

\begin{table}[htbp]
	\renewcommand{\arraystretch}{1.2}
	\caption{Details of Datasets.}
	\label{table1}
	\centering
	\renewcommand\tabcolsep{3.0pt}
	\begin{tabular}{lcccccc}
		\toprule[1pt]
	{Datasets} &{Rename} &{Samples} & {Features}& {Classes} &{IR} &{Source}
        \\ \hline
     Credit Approval   &S1      &690     &15   &2     &1.25  &\cite{6}\\
     Diabetes          &S2      &768     &8    &2     &1.87  &\cite{6}\\
     Car Evaluation    &S3      &1728    &6    &4     &18.62 &\cite{6}\\
     Pumpkin Seeds     &S4      &2500    &12   &2     &1.08  &\cite{49}\\
     banana            &S5      &5300    &2    &2     &1.23  &\cite{7}\\
     page-blocks       &S6      &5473    &11   &5     &175.46 &\cite{6}\\
     coil2000          &S7      &9822    &85   &2     &15.76 &\cite{7}\\
     Dry Bean          &S8      &13611   &16   &7     &6.79   &\cite{6}   \\
     HTRU2             &S9      &17898   &8    &2     &9.92 &\cite{6}\\
     magic             &S10     &19020   &10   &2     &1.84  &\cite{7}\\
     shuttle           &S11     &58000   &9    &7     &4558.6  &\cite{7}\\
     Gas Sensor        &S12     &13910   &128  &6     &1.83   &\cite{6}\\
     USPS              &S13     &9298    &256  &10    &2.19  &\cite{50}\\
		\toprule[1pt]
	\end{tabular}
\vspace{-0.2in}
\end{table}

\subsection{Analysis of Sampling Ratio}
\label{AnalofCR}
This section analyzes and discusses the performance of GBABS in data compression on standard datasets and class noise datasets.

\begin{figure}[htbp]
	\centering
	\begin{minipage}[t]{0.5\linewidth}
		\centering
		\includegraphics[height=3.3cm, width=4.4cm]{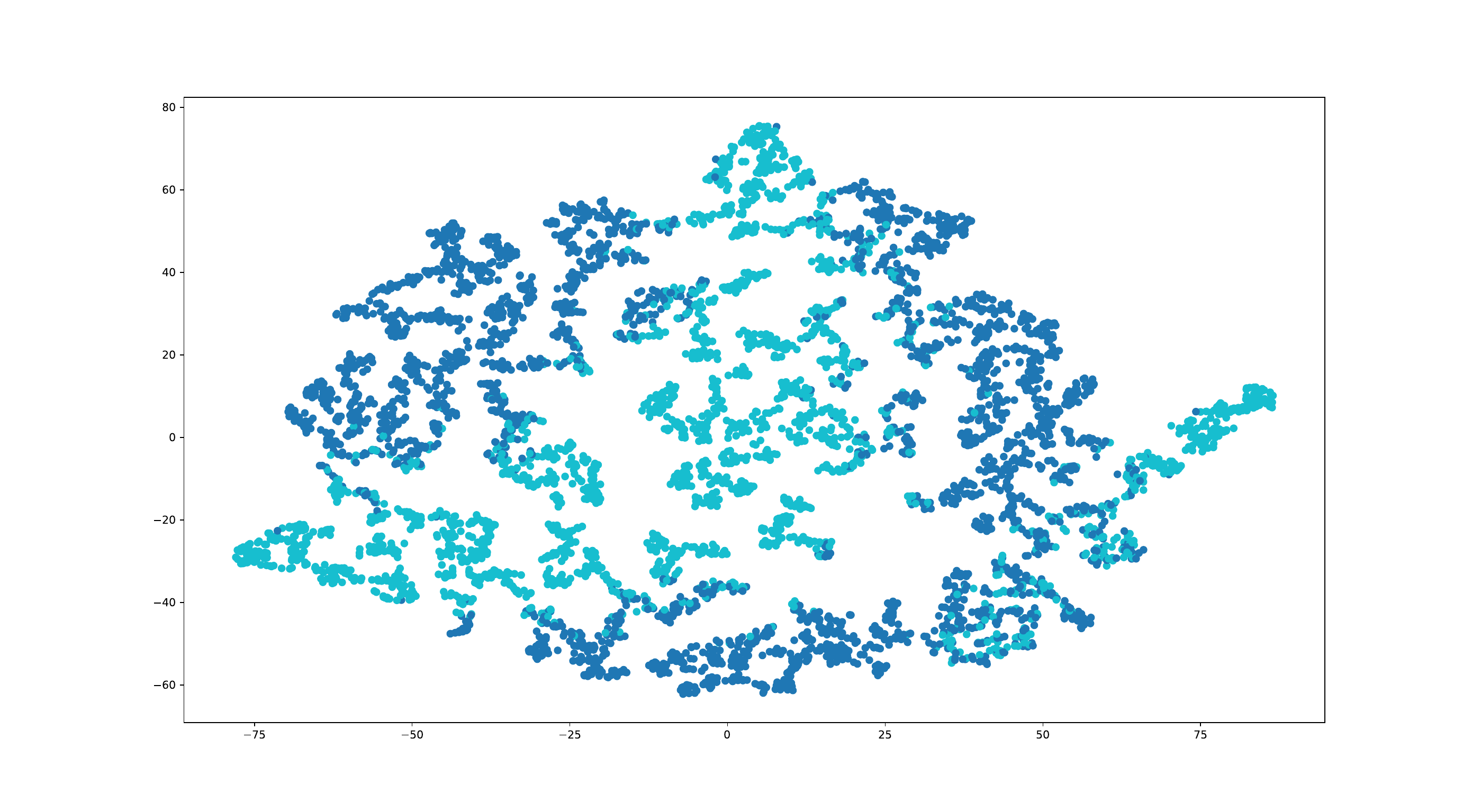}
		\small (a) Dataset S5.
	\end{minipage}%
	\begin{minipage}[t]{0.5\linewidth}
		\centering
		\includegraphics[height=3.3cm, width=4.4cm]{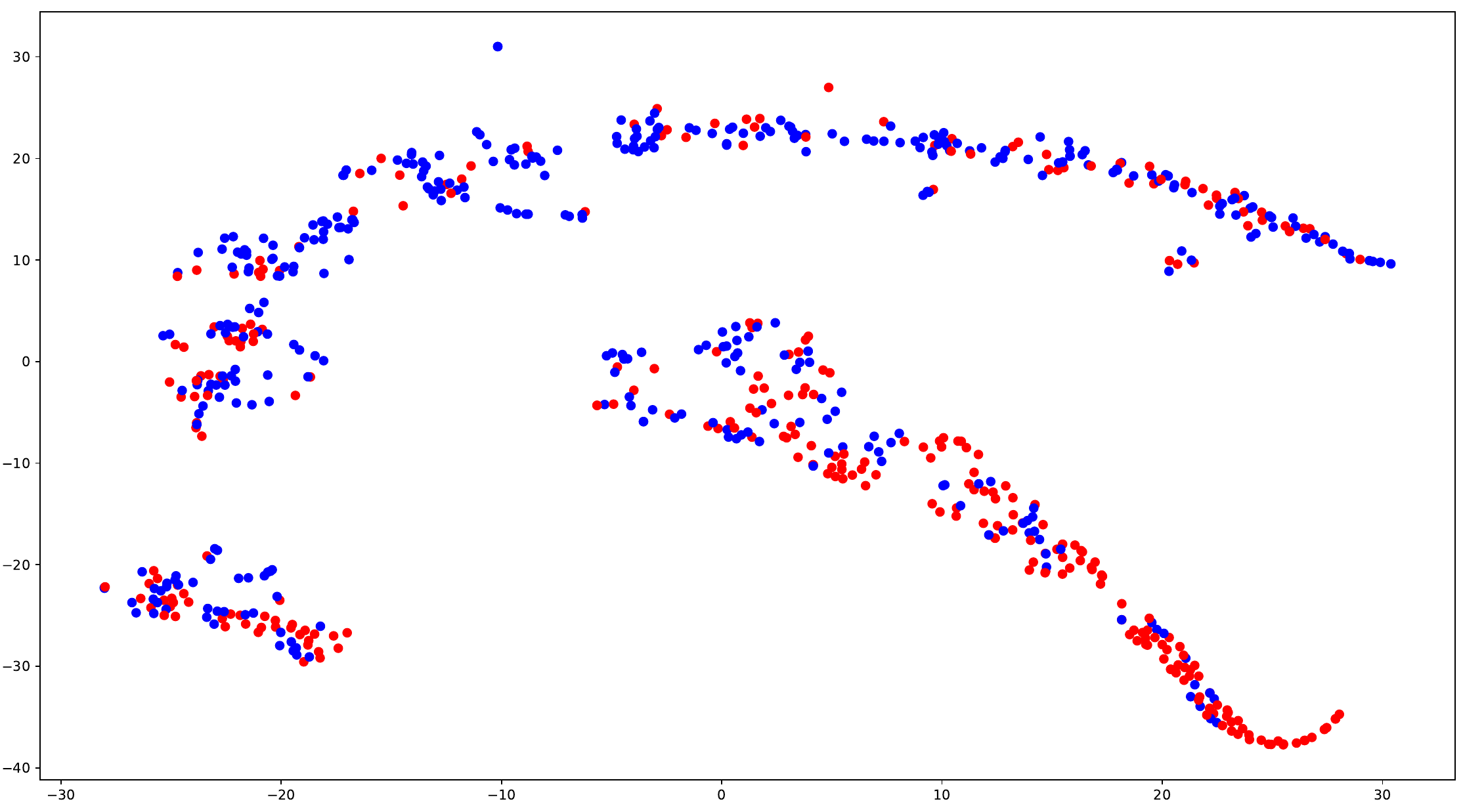}
		\small (b) Dataset S1.
	\end{minipage}%

	\begin{minipage}[t]{0.5\linewidth}
		\centering
		\includegraphics[height=3.3cm, width=4.4cm]{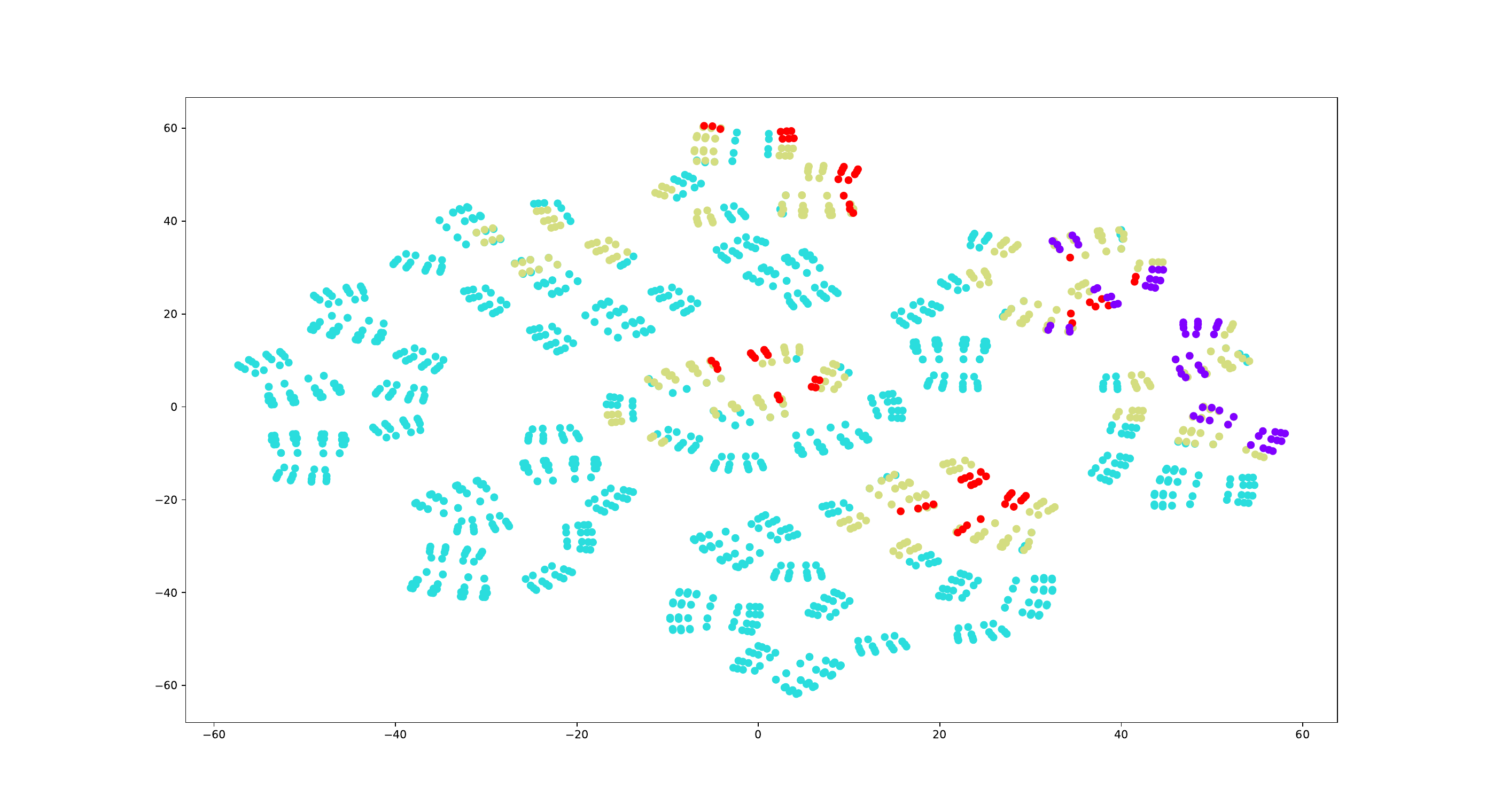}
		\small (c) Dataset S3.
	\end{minipage}%
	\begin{minipage}[t]{0.5\linewidth}
		\centering
		\includegraphics[height=3.3cm, width=4.4cm]{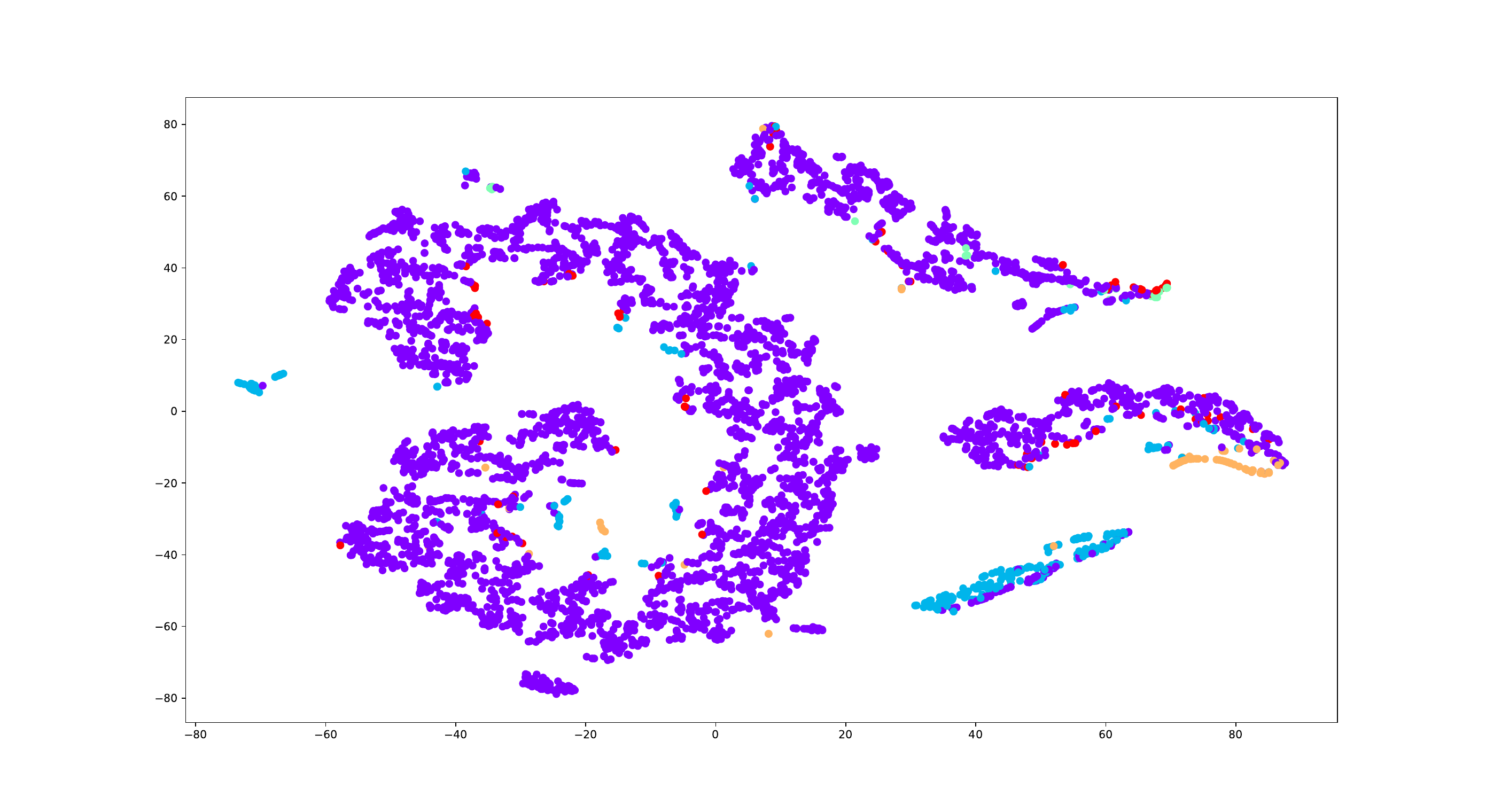}
		\small (d) Dataset S6.
	\end{minipage}%
\caption {Visualization of several datasets. }
\label{fig5}
\end{figure}

\begin{figure*}[htbp]
\vspace{-0.5cm}
\centering
        \begin{minipage}[t]{0.45\linewidth}
        \centering
        \includegraphics[height=1.5in,width=2.5in]{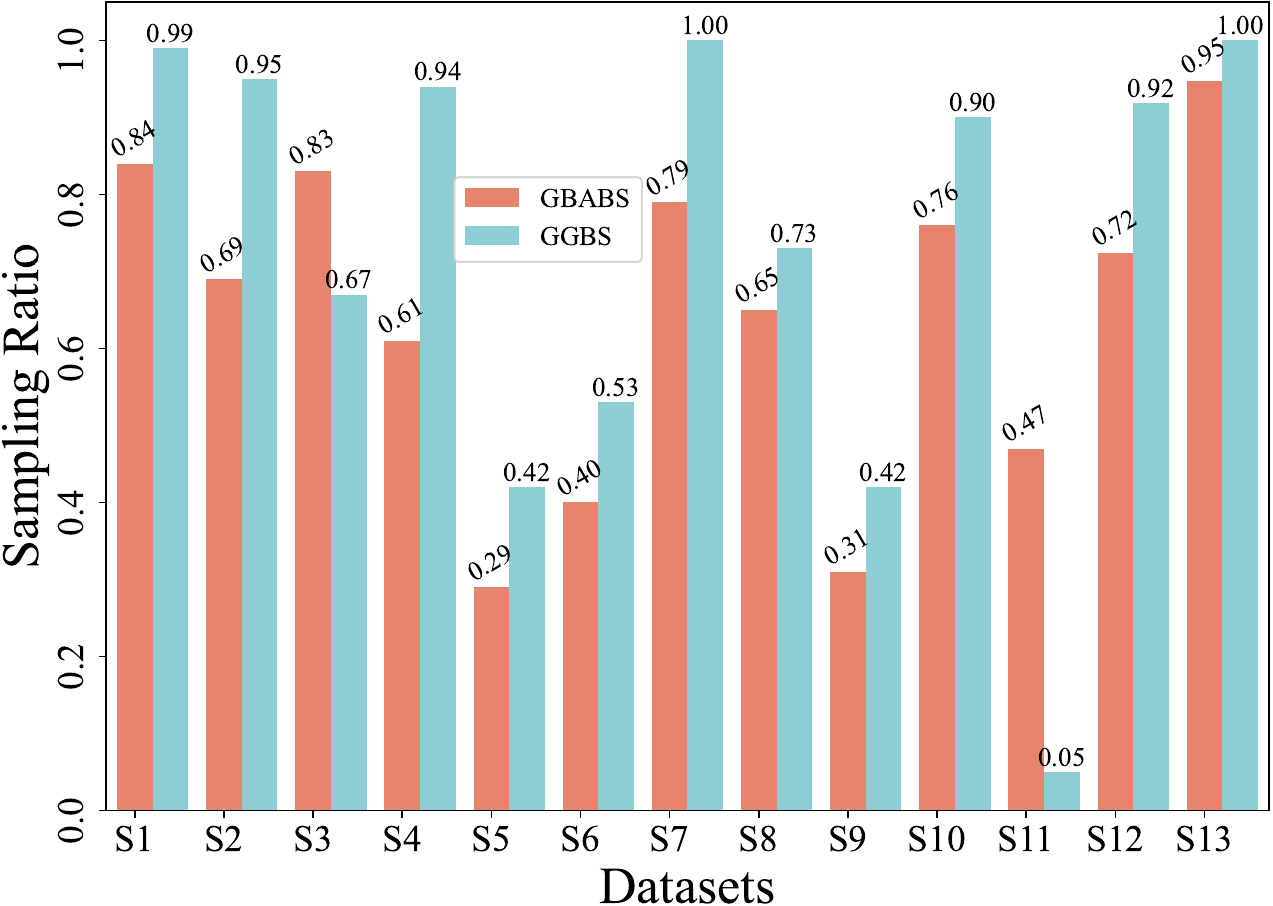}

        \small (a) Noise ratio: $0\%$.
        \end{minipage}%
        \begin{minipage}[t]{0.45\linewidth}
        \centering
        \includegraphics[height=1.5in,width=2.5in]{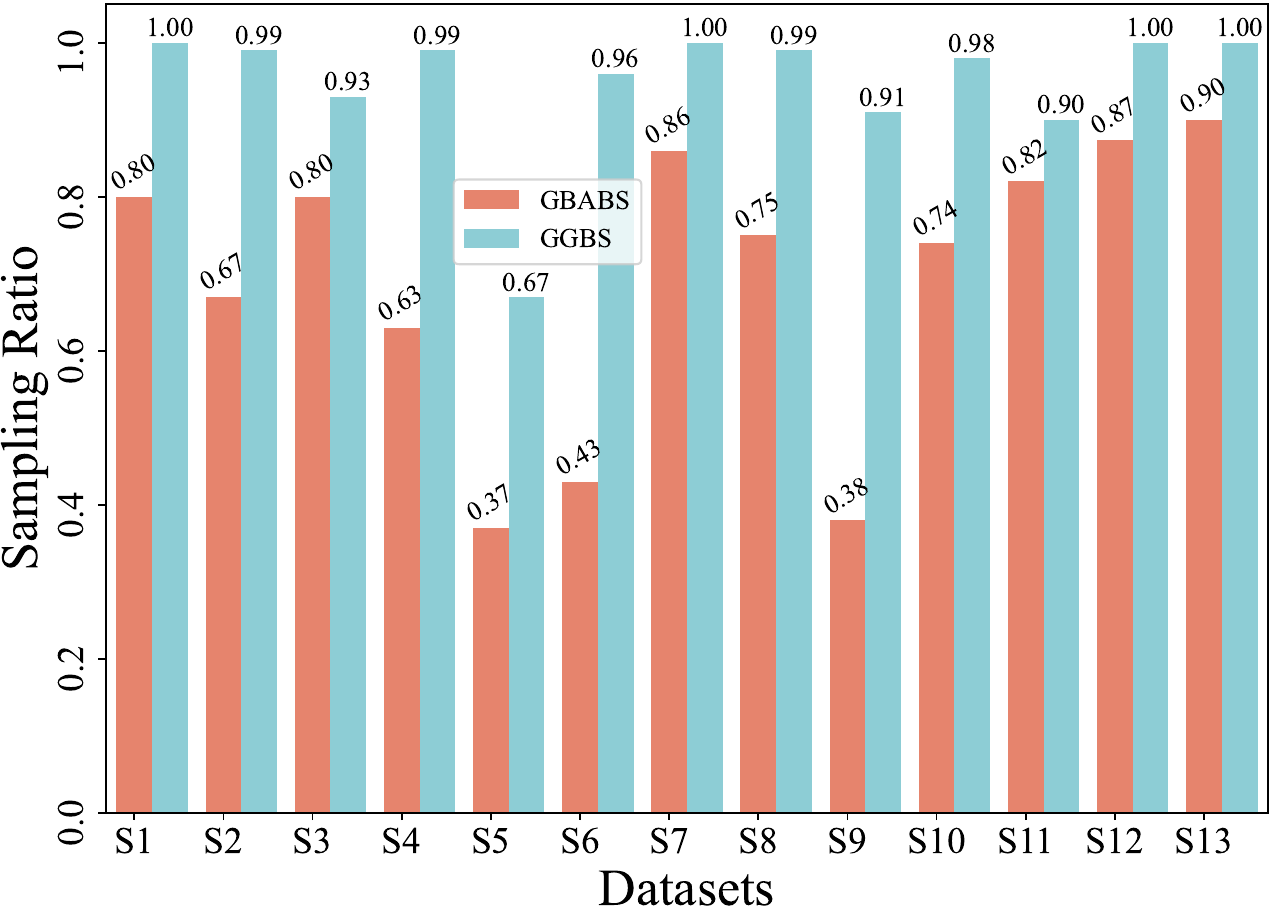}

        \small (b) Noise ratio: $5\%$.
        \end{minipage}%
        \vspace{0.1in}
\quad
        \begin{minipage}[t]{0.45\linewidth}
        \centering
        \includegraphics[height=1.5in,width=2.5in]{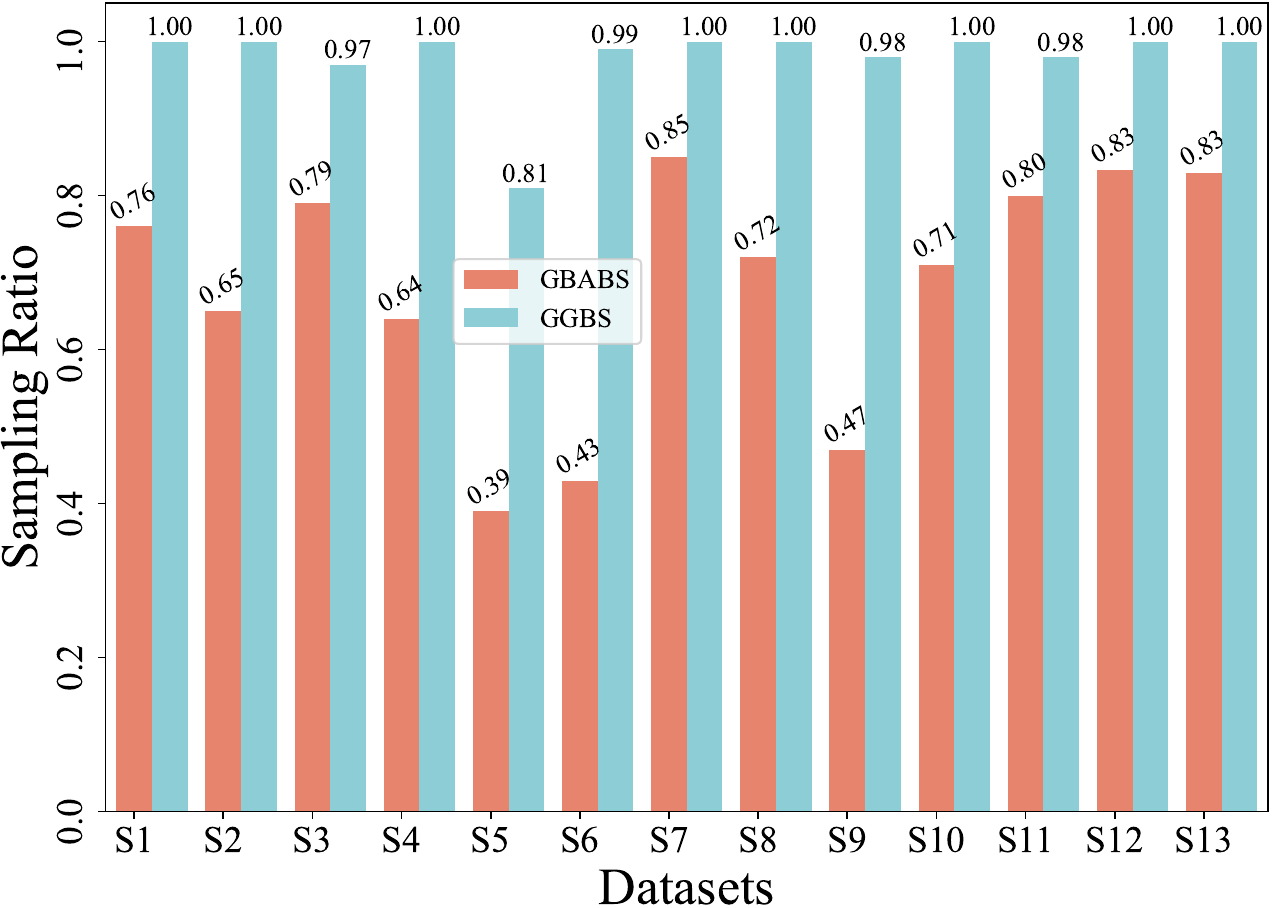}

        \small (c) Noise ratio: $10\%$.
        \end{minipage}%
        \begin{minipage}[t]{0.45\linewidth}
        \centering
        \includegraphics[height=1.5in,width=2.5in]{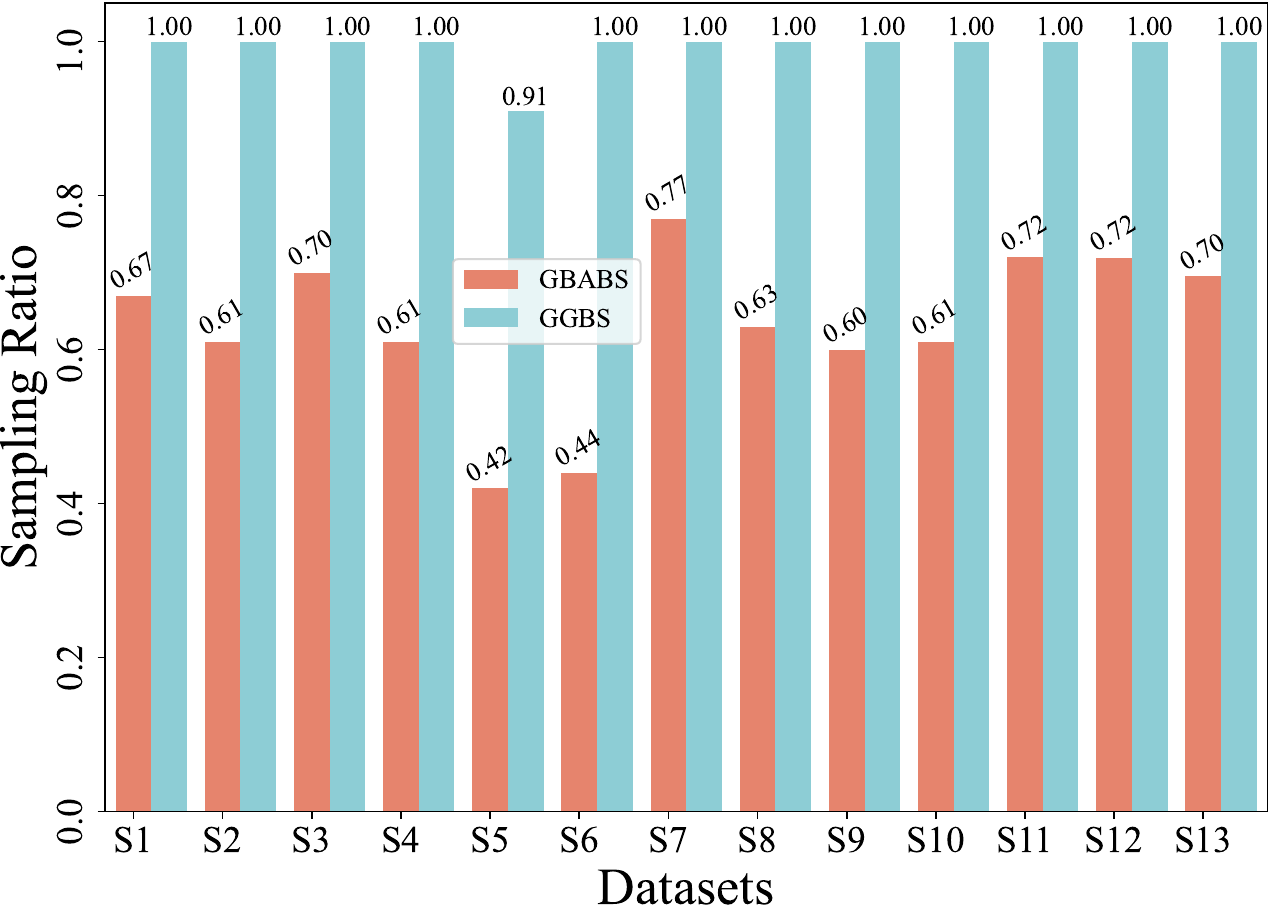}

        \small (d) Noise ratio: $20\%$.
        \end{minipage}%
        \vspace{0.1in}

\quad
        \begin{minipage}[t]{0.45\linewidth}
        \centering
        \includegraphics[height=1.5in,width=2.5in]{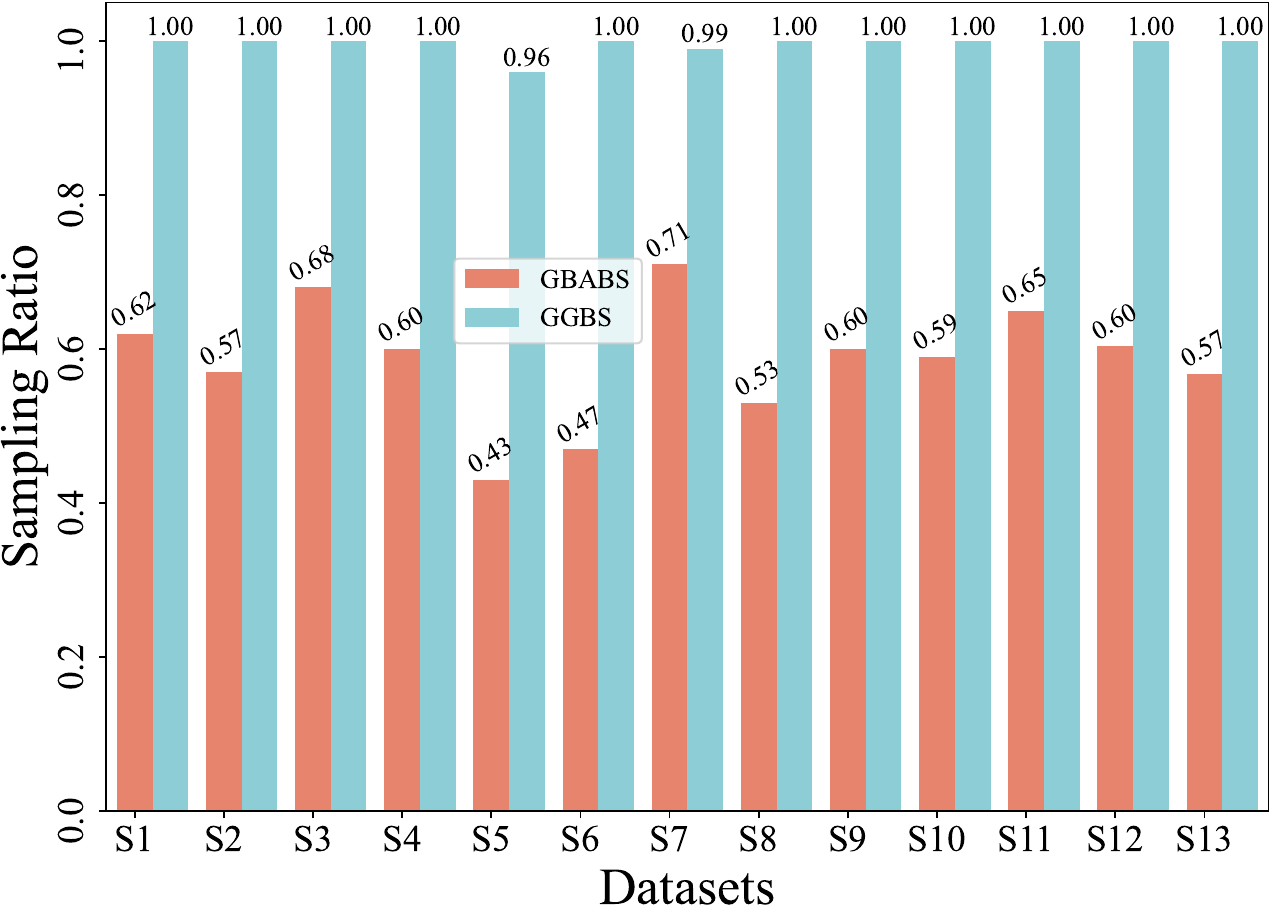}

        \small (e) Noise ratio: $30\%$.
        \end{minipage}%
        \begin{minipage}[t]{0.45\linewidth}
        \centering
        \includegraphics[height=1.5in,width=2.5in]{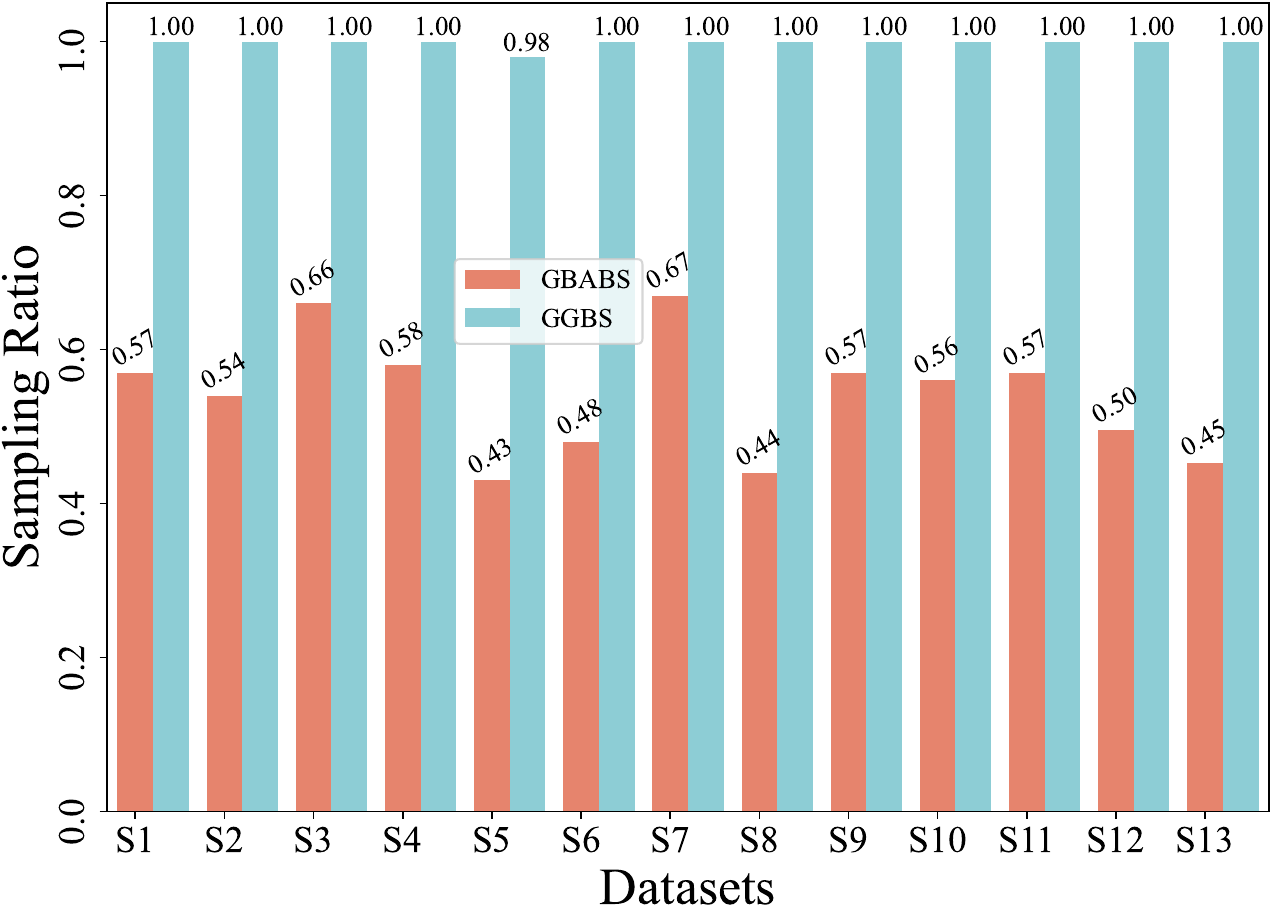}

        \small (f) Noise ratio: $40\%$.
        \end{minipage}%

  \caption {Comparison of sampling ratio under different class noise ratios.}
\label{fig6}
\end{figure*}

Fig. \ref{fig6}(a) provides insights into the sampling ratio of GBABS and GGBS on each standard dataset listed in Table\ref{table1}. Additionally, Fig.\ref{fig5} visualizes several standard datasets using TSNE, a dimensionality reduction technique, with different classes marked with different colors. Observation from Fig.\ref{fig6}(a) reveals that GBABS achieves notable compression across all datasets, with a minimum sampling ratio of approximately $29\%$. The reason why GBABS has excellent data compression capability is that GBABS samples on GBs, and the number of GBs is generally much smaller than the sample size of the original dataset.
Notably, GBABS exhibits a smaller sampling ratio on datasets with lower dimensions or fewer classes. For example, the sampling ratio for the two-dimensional binary dataset S5 is about $29\%$, while for the higher-dimensional dataset S1, the ratio increases to approximately $84\%$. Fig. \ref{fig5}(a) and (b) illustrate that the class boundaries of S5 are relatively simple, in contrast to the complex boundaries of S1 due to its high dimensionality. In high-dimensional spaces, retaining more samples is crucial due to the increased complexity of class boundaries.

Besides, as depicted in Fig.\ref{fig6}(a), GBABS generally exhibits lower sampling ratios compared to GGBS across most datasets, such as datasets S6 and S10. Notably, for a high-dimensional dataset such as dataset S7, the sampling ratio of GGBS is $1.0$, which means that the sampling capability of GGBS is invalid. For some datasets with unclear class boundaries, the sampling ratio of GBABS is slightly higher than that of GGBS, such as the dataset S3. As shown in Fig.\ref{fig5}(c), the distributions of samples of different classes in S3 overlap in the feature space, so the class boundaries are unclear. However, retaining sufficient borderline samples for high-dimensional datasets or datasets with complex and blurred class boundaries is crucial to ensure effective classification. Besides, on a dataset with relatively clear class boundaries, even if it is multi-class, such as a dataset S6 whose visualization is shown in Fig.\ref{fig5}(d), GBABS yields a smaller scale sampled dataset than GGBS. Generally, the superior data compression capability of GBABS is attributed to its selective sampling on borderline GBs, unlike GGBS, which samples on each GB.

Furthermore, Fig. \ref{fig6}(b)-(f) depict the sampling ratios of GBABS and GGBS on datasets with class noise ratios of $5\%$, $10\%$, $20\%$, $30\%$, and $40\%$, respectively. It can be observed that under any noise ratio, the data sampling ratio of GBABS is always lower than that of GGBS. Specifically, on dataset S8 at $20\%$ noise ratio, GBABS achieves a sampling ratio as low as $44\%$, whereas GGBS retains a $100\%$ sampling ratio. Compared to standard datasets, GBABS exhibits stronger data compression on class noise datasets, with its advantage over GGBS becoming more pronounced as the noise ratio increases.

Two factors contribute to GBABS's superior data compression on class noise datasets. First, the RD-GBG method (Section \ref{RDA-GBG}) incorporates noise elimination, removing most class noise samples as the noise ratio increases, while the GBG method of GGBS does not consider that. Second, GBABS focuses on sampling from borderline GBs, unlike GGBS, which samples uniformly from all GBs. This results in a lower sampling ratio for GBABS, even when class noise requires more GBs to cover.

\subsection{Effectiveness on Standard Datasets}
\label{EffonSD}
This section mainly validates the lossless compression capability of GBABS as a sampling method tailored for classification tasks.
Table \ref{table2} shows the testing $Accuracy$ for the GBABS-based DT, GGBS-based DT, SRS-based DT, and DT on standard datasets listed in Table \ref{table1}.
Table \ref{table3} reports the results of the Wilcoxon signed-rank test for the comparison between GBABS-based DT and others. The results indicate significant differences in performance across all tested pairs at the $0.05$ significance level, which strongly suggests that the GBABS-based DT consistently outperforms the others across all comparisons.

\begin{table}[htbp]
	\renewcommand{\arraystretch}{1.2}
	\caption{Comparison on testing $Accuracy$ of DT with different sampling methods.}
	\label{table2}
	\centering
	\renewcommand\tabcolsep{3.6pt}
	\begin{tabular}{lcccc}
		\toprule[1pt]
		{Datasets} &{GBABS-DT} & {GGBS-DT}& {SRS-DT} & {DT}
        \\ \hline
     S1     &\textbf{0.8577} 	&0.8145 	&0.7968 	&0.8145 \\
     S2   	&\textbf{0.7351} 	&0.6936 	&0.6825 	&0.6902 \\
     S3     &\textbf{0.8851} 	&0.8737 	&0.8763 	&0.8744 \\
     S4     &\textbf{0.8721} 	&0.8338 	&0.8345 	&0.8344 \\
     S5     &0.8709 	&0.8528 	&0.8638 	&\textbf{0.8728} \\
     S6   	&\textbf{0.9667} 	&0.9606 	&0.9592 	&0.9646 \\
     S7   	&\textbf{0.9348} 	&0.8969 	&0.8913 	&0.8965 \\
     S8   	&\textbf{0.9009} 	&0.8892 	&0.8925 	&0.8950 \\
     S9     &\textbf{0.9761} 	&0.9576 	&0.9662 	&0.9680 \\
     S10    &\textbf{0.8396} 	&0.8152 	&0.8152 	&0.8129 \\
     S11    &0.9994 	&0.9983 	&0.9995 	&\textbf{0.9998} \\
     S12   	&0.9693     &0.9684  &0.9675  &\textbf{0.9750}  \\
     S13    &\textbf{0.8846}     &0.8843  &0.8826  &0.8843  \\
        \hline
        Average     &\textbf{0.8994} 	&0.8799 	&0.8791 	&0.8832 \\
		\toprule[1pt]
	\end{tabular}
\end{table}

\begin{table}[htbp]
	\renewcommand{\arraystretch}{1.0}
	\caption{Wilcoxon signed-rank test results.}
	\label{table3}
	\centering
	\renewcommand\tabcolsep{3.6pt}
	\begin{tabular}{lccc}
		\toprule[1pt]
		{Comparison Method} & {p-value}& {Significance ($\alpha = 0.05$)}
        \\ \hline
    GBABS-DT vs. GGBS-DT		&0.000244	&Significant  \\
    GBABS-DT vs. SRS-DT	   	&0.000488	&Significant   \\
    GBABS-DT vs. DT	        	&0.010498	&Significant  \\
		\toprule[1pt]
	\end{tabular}
\end{table}

Specifically, GBABS-based DT outperforms DT on $77\%$ of the datasets, such as the testing $Accuracy$ on dataset S2 is improved by $0.0449$. This superior performance stems from the approximate borderline sampling strategy detailed in Section \ref{GBABS} ensures the collection of samples crucial for delineating class boundaries, thereby enhancing the ability of the sampled dataset to describe the class boundary of the original dataset. Furthermore, by abstaining from collecting intra-class samples, GBABS mitigates overfitting and the impact of noisy data to a significant extent.

Compared with GGBS-based DT, it's evident that GBABS-based DT consistently achieves higher testing $Accuracy$. This disparity can be attributed to several factors. First, GGBS refrains from splitting the GB with a sample size less than or equal to twice the number of features, irrespective of purity thresholds, which diminishes the quality of generated GBs, thus impairing their effectiveness in describing the original dataset. Second, the overlap between GBs in the GBG method of GGBS can result in blurred class boundaries, undermining classification performance. Third, the center selection method (introduced in Section \ref{DMLDC}) and the radius determination rule of GB (introduced in Section \ref{GBG}) enable the GBs constructed by RD-GBG to more accurately represent the original dataset compared to those constructed by the existing GBG method. Fourth, the GBABS aims to collect samples near the class boundaries, where the borderline samples are crucial for classification. In contrast, GGBS samples from all GBs, including redundant samples, may degrade classifier performance.

The testing $Accuracy$ of GBABS-based DT is higher than that of SRS-based DT on almost all datasets. The reason is that GBABS is essentially a biased sampling method, and samples on the class boundary have a higher probability of being sampled, while SRS is an unbiased sampling method. Consequently, when SRS and GBABS adopt the same sampling ratio for a given dataset, GBABS retains more borderline samples, enriching the sampled dataset with more effective information for classifiers.

\subsection{Robustness to Class Noise Datasets}
\label{RobustoCND}
This section mainly verifies the enhancement of the robustness of the classifier by GBABS. Considering that different classifiers have different sensitivity to class noise, five commonly used and representative machine learning classifiers are employed to obtain comprehensive and reliable results and alleviate the bias of comparative experimental settings. Comparative experiments are conducted on datasets with class noise ratios of $5\%$, $10\%$, $20\%$, $30\%$, and $40\%$, respectively.

\begin{table}[htbp]
	\renewcommand{\arraystretch}{1.2}
	\caption{Comparison on average testing $Accuracy$ on class noise datasets.}
	\label{table4}
	\centering
	\renewcommand\tabcolsep{3.6pt}
	\begin{tabular}{lccccc}
		\toprule[1pt]
		{Noise ratio} &{$5\%$} &{$10\%$} & {$20\%$}& {$30\%$} & {$40\%$}
        \\ \hline
        GBABS-DT       &\textbf{0.8598}  	&\textbf{0.8004} 	&\textbf{0.6955} 	&\textbf{0.5991} 	&\textbf{0.5133} \\
        GGBS-DT         &0.8063            &0.7206           &0.6036 	        &0.5126 	        &0.4433 \\
        SRS-DT         &0.8079            &0.7239           &0.5998 	        &0.5109 	        &0.4409 \\
        DT             &0.8097            &0.7239           &0.6037 	        &0.5126 	        &0.4431 \\
        \hline
        GBABS-XGBoost  &\textbf{0.8719}     &\textbf{0.8243}   &\textbf{0.7325}    &\textbf{0.6384}   &\textbf{0.5449}  \\
        GGBS--XGBoost   &0.8658              &0.8165            &0.7155             &0.6200            &0.5295           \\
        SRS--XGBoost   &0.8643              &0.8126            &0.7106             &0.6100            &0.5206           \\
        XGBoost        &0.8673              &0.8170            &0.7155             &0.6200            &0.5293           \\
        \hline
        GBABS-LightGBM &0.8660            &0.8166            &\textbf{0.7338}    &\textbf{0.6422}    &\textbf{0.5515}  \\
        GGBS-LightGBM   &\textbf{0.8690}  &0.8219            &0.7285             &0.6359             &0.5414           \\
        SRS-LightGBM   &0.8669            &0.8184            &0.7203             &0.6257             &0.5303           \\
        LightGBM       &0.8685            &\textbf{0.8222}    &0.7281             &0.6361             &0.5416           \\
        \hline
        GBABS-$k$NN	   &\textbf{0.8642}   &\textbf{0.8213}      &\textbf{0.7262}   &\textbf{0.6315}  &\textbf{0.5432} \\
        GGBS-$k$NN	     &0.8633            &0.8155               &0.7138            &0.6096           &0.5173          \\
        SRS-$k$NN	     &0.8622            &0.8141               &0.7089            &0.6061           &0.5158          \\
        $k$NN	         &0.8636            &0.8159               &0.7143            &0.6097           &0.5177          \\
        \hline
        GBABS-RF	     &\textbf{0.8732} &\textbf{0.8277}  &\textbf{0.7340}   &\textbf{0.6430}   &\textbf{0.5501}  \\
        GGBS-RF	       &0.8693          &0.8194           &0.7211            &0.6199            &0.5253           \\
        SRS-RF	       &0.8693          &0.8200           &0.7183            &0.6193            &0.5250           \\
        RF	           &0.8698          &0.8203           &0.7206            &0.6196            &0.5246           \\
		\toprule[1pt]
	\end{tabular}
\vspace{-0.2cm}
\end{table}

Table \ref{table4} shows the average testing $Accuracy$ of GBABS-based classifier, GGBS-based classifier, SRS-based classifier, and classifier on datasets with different noise ratios, where classifiers are DT, XGBoost, lightgbm, RF, and $k$NN.
It can be observed from Table \ref{table4} that the GBABS-based classifier generally performs better, whether in the case of low noise ratio (such as $5\%$) or high noise ratio (such as $40\%$). Especially in high-noise environments, GBABS can maintain a relatively stable performance for each classifier compared with others.

\begin{figure}[htbp]
\centering
\includegraphics[height=2.0in,width=3.3in]{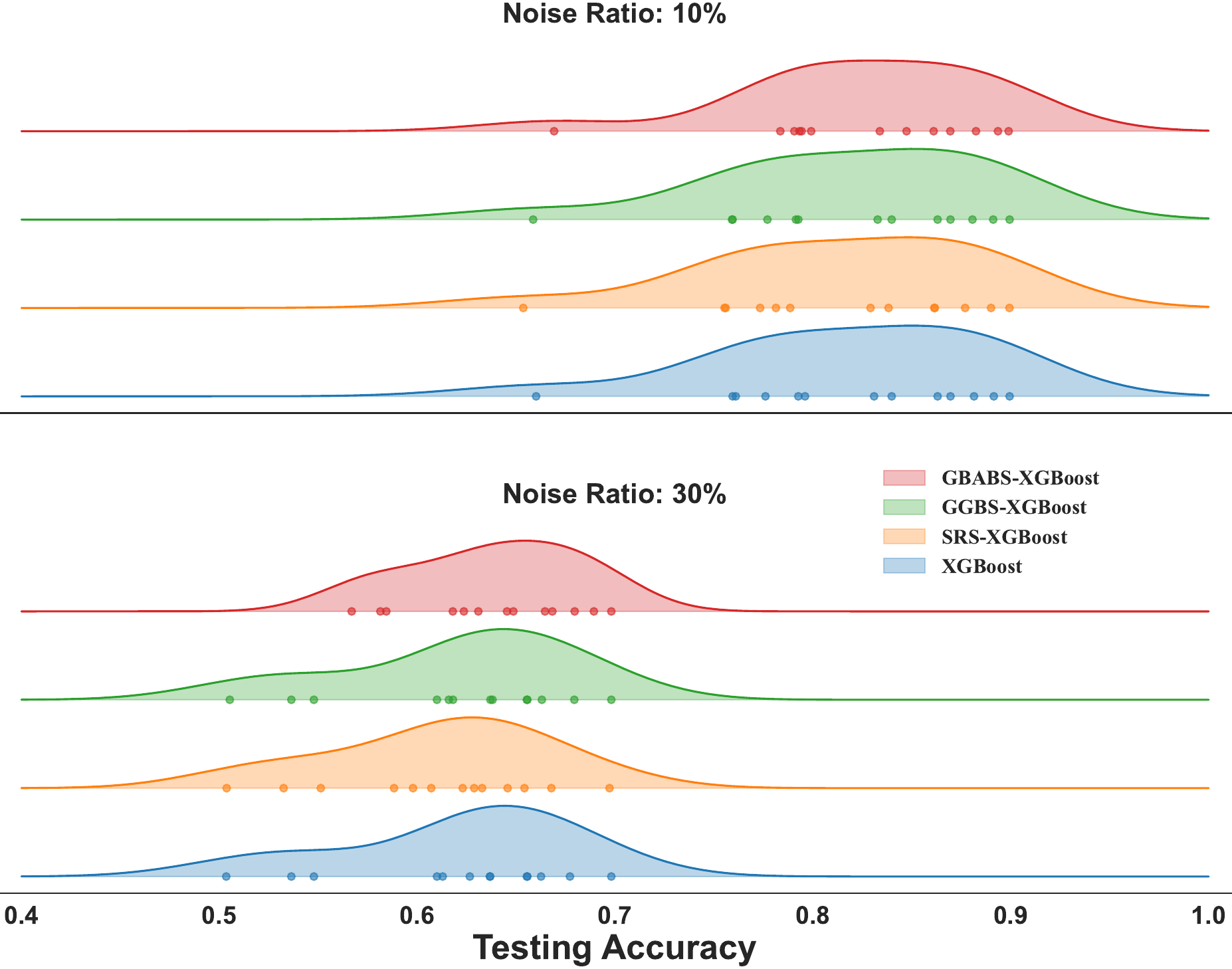}
\caption {Distribution of testing $Accuracy$ for XGBoost with different sampling methods at different noise ratios.}
\label{fig10}
\end{figure}

\begin{figure}[htbp]
\centering
\includegraphics[height=2.0in,width=3.3in]{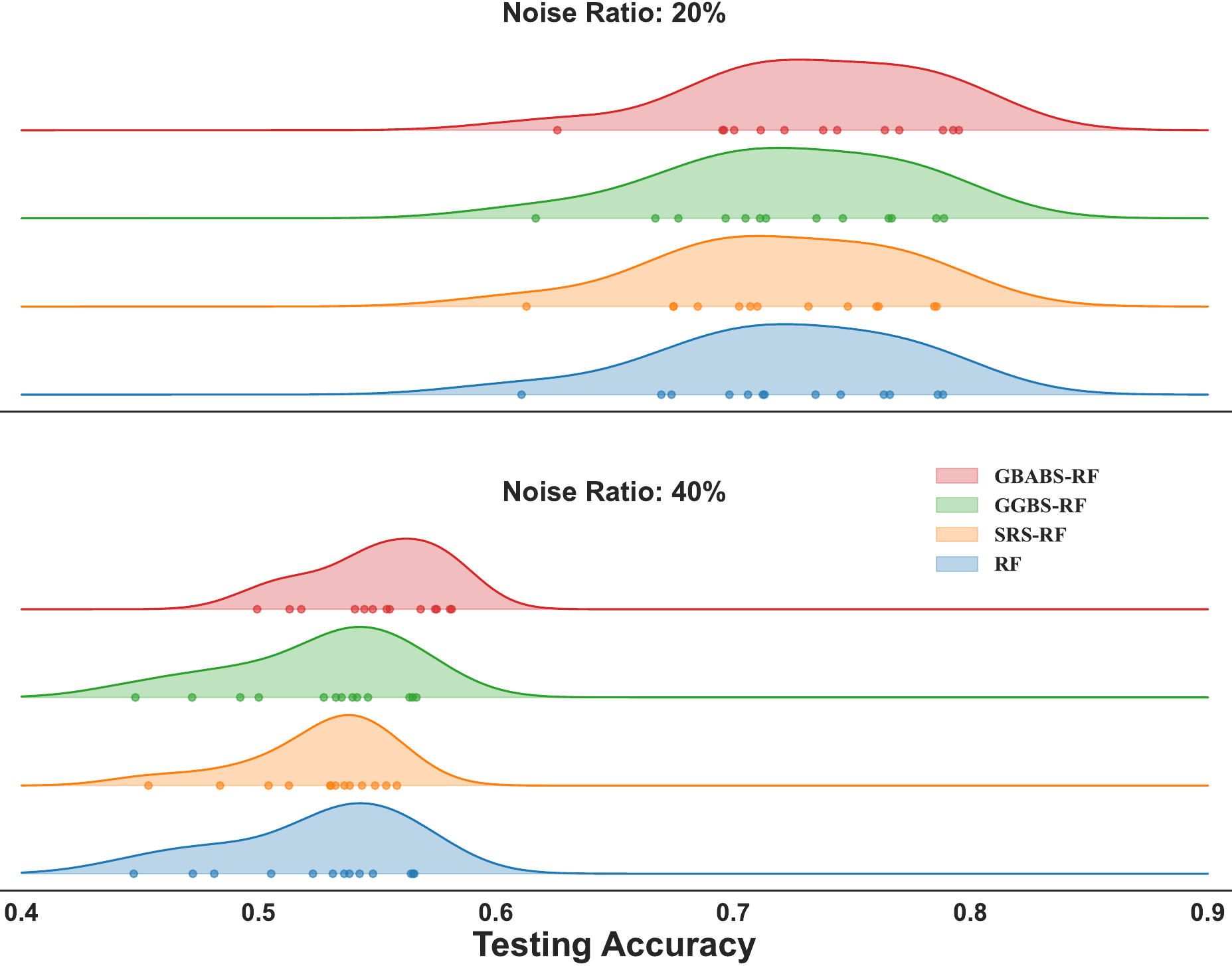}
\caption {Distribution of testing $Accuracy$ for RF with different sampling methods at different noise ratios.}
\label{fig11}
\end{figure}

The ridge plot shown in Fig. \ref{fig10} shows the distribution of testing $Accuracy$ for XGBoost with different sampling methods at noise ratios of $10\%$ and $30\%$, while Fig. \ref{fig11} presents the distribution for RF at noise ratios of $20\%$ and $40\%$. Curves of different colors represent the $Accuracy$ distribution of different sampling methods. The scatter points of different colors represent the testing $Accuracy$ of different methods on each dataset. From the distribution in the ridge plot, it can be seen that under different noise conditions, whether used for XGboost or RF, GBABS shows higher consistency and stability. Especially when the noise ratio is high (such as $40\%$), the testing $Accuracy$ distribution of GBABS-based RF shifts to the right (peak value is about $0.55-0.6$), which is significantly better than other methods. When the noise ratio is low (e.g., $10\%$), the testing $Accuracy$ distribution of the GBABS-based classifier is concentrated, and the peak value is slightly higher than that of others.

In conclusion, it can be inferred that GBABS can effectively enhance the robustness of classifiers, outperforming GGBS and SRS. There are two primary reasons why GBABS excels in enhancing robustness. First, the RD-GBG (introduced in Section \ref{RDA-GBG}) considers noise elimination, whereas GGBS and SRS do not. Second, GBABS (introduced in Section \ref{GBABS}) is designed to collect samples on class boundaries, thereby avoiding the collection of redundant and noisy samples. In contrast, GGBS and SRS also do not consider that.

\subsection{Effectiveness on Imbalanced Datasets}
\label{EffonID}

\begin{figure}[htbp]
	\centering
	\begin{minipage}[t]{0.45\linewidth}
		\centering
		\includegraphics[height=3.0cm, width=4.2cm]{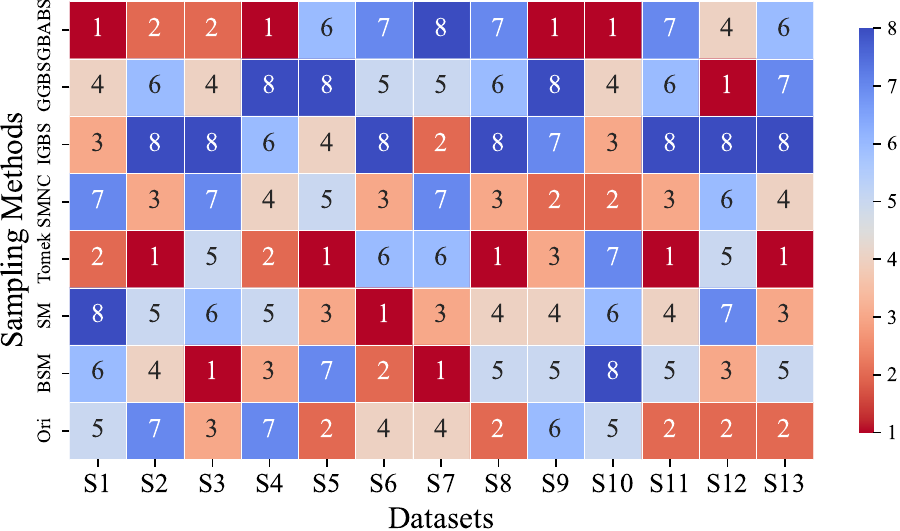}
		\small (a) Noise ratio: $0\%$.
	\end{minipage}%
        \hspace{0.04\linewidth} 
	\begin{minipage}[t]{0.45\linewidth}
		\centering
		\includegraphics[height=3.0cm, width=4.2cm]{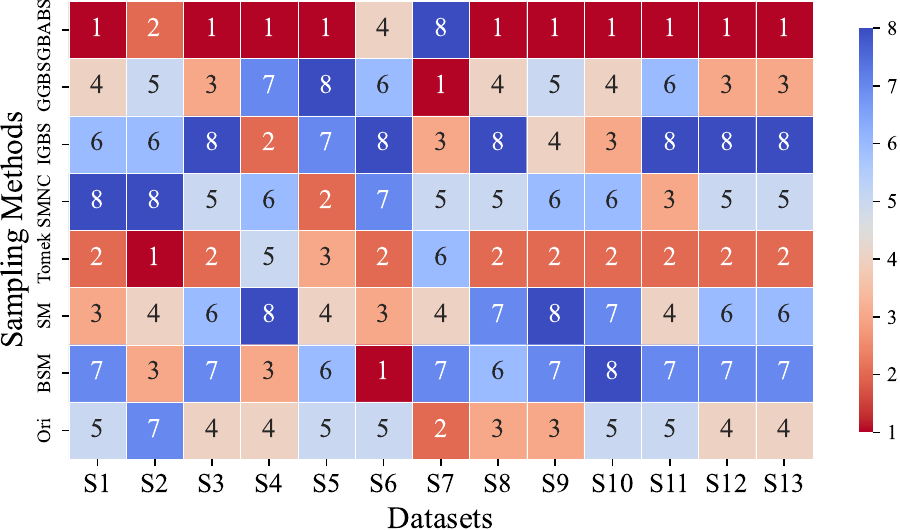}
		\small (b) Noise ratio: $5\%$.
	\end{minipage}%

	\begin{minipage}[t]{0.45\linewidth}
		\centering
		\includegraphics[height=3.0cm, width=4.2cm]{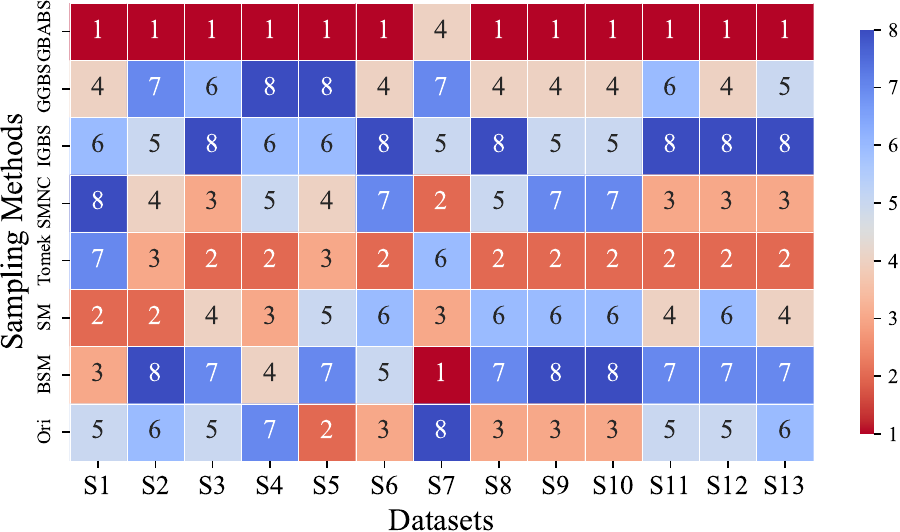}
		\small (c) Noise ratio: $10\%$.
	\end{minipage}%
        \hspace{0.04\linewidth} 
	\begin{minipage}[t]{0.45\linewidth}
		\centering
		\includegraphics[height=3.0cm, width=4.2cm]{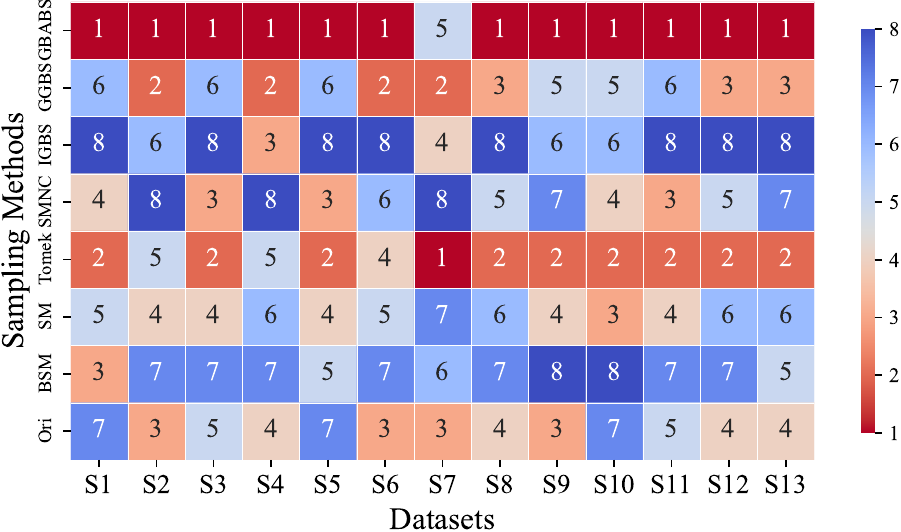}
		\small (d) Noise ratio: $20\%$.
	\end{minipage}%

        \begin{minipage}[t]{0.45\linewidth}
        	\centering
        	\includegraphics[height=3.0cm, width=4.2cm]{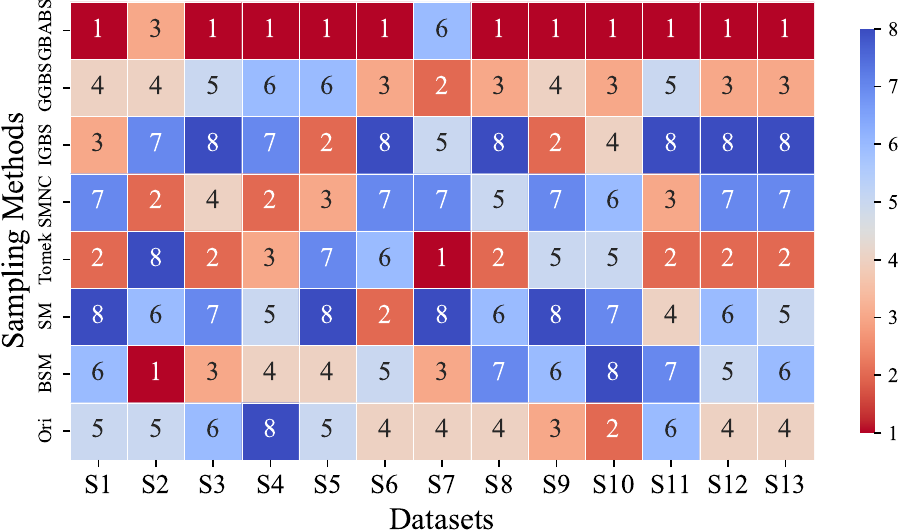}
        	\small (e) Noise ratio: $30\%$.
        \end{minipage}%
        \hspace{0.04\linewidth} 
        \begin{minipage}[t]{0.45\linewidth}
        	\centering
        	\includegraphics[height=3.0cm, width=4.2cm]{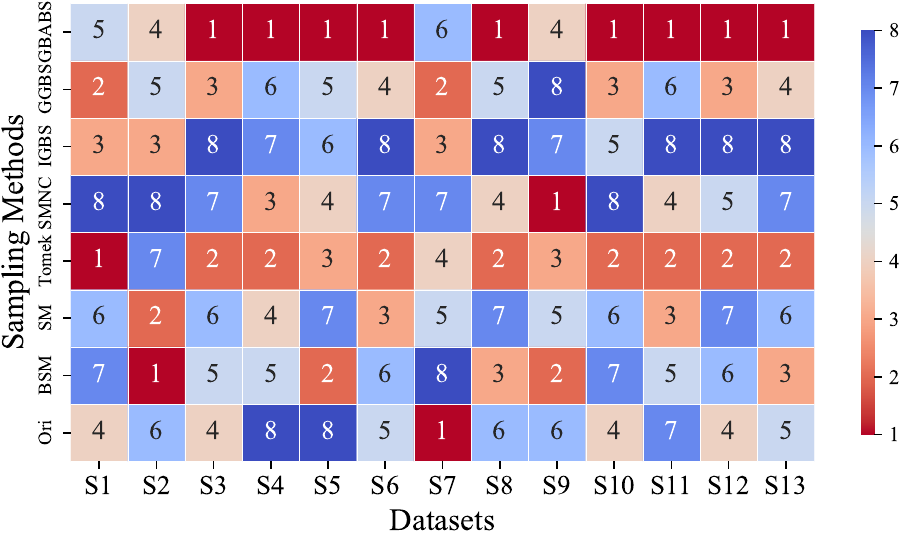}
        	\small (f) Noise ratio: $40\%$.
        \end{minipage}%
\caption {Comparison on ranking of testing $G-mean$ for DT with various sampling methods at each noise ratio.}
\label{fig7}
\end{figure}

This section mainly evaluates the effectiveness of GBABS in mitigating the bias problem of standard imbalanced datasets (including binary and multi-class datasets) and imbalanced datasets with class noise. Fig.\ref{fig7}(a) demonstrates the ranking of testing $G-mean$ of DT on the standard datasets when GBABS, GGBS, IGBS, SM, BSM, SMNC, and Tomek are used as the sampling methods, while Fig.\ref{fig7}(b)-(f) shows the ranking of testing $G-mean$ on datasets with class noise ratios of $5\%$, $10\%$, $20\%$, $30\%$, and $40\%$, respectively. The larger the value of $G-mean$, the higher the ranking. It can be observed that, on most standard imbalanced datasets, GBABS-based DT ranks high, and on almost all imbalanced datasets with class noise, GBABS-based DT achieves the best performance. In a high noise environment (such as $40\%$), although the ranking of GBABS has dropped on a few datasets, it is still better than most other sampling methods. The reason is that, as seen in Fig. \ref{fig6}, as the ratio of class noise increases, the sampling ratio of GBABS on each dataset decreases. Too few samples may reduce performance for datasets with small sample sizes, such as S1 and S2.
In conclusion, GBABS can mitigate the bias issue caused by class imbalance to a certain extent, especially performing excellently in scenarios with class noise.

There are three main reasons why GBABS exhibits superior performance in handling imbalanced datasets. First, GBABS is essentially an undersampling method that aims to sample borderline samples. Therefore, the removal of redundant samples in the majority class is generally more aggressive than that in the minority class, which can alleviate the class imbalance to a certain extent and reduce the overfitting of the classifier to majority class samples. Additionally, as mentioned in Section \ref{AnalofCR}, when the dimensionality of the dataset is high, or the dataset size is small, the compression ability of GGBS and IGBS may fail, i.e., they cannot effectively undersample the majority class samples. Second, compared with oversampling methods such as SM, BSM, and SMNC, GBABS avoids the risks of synthetic samples, such as introducing noise or overfitting, particularly in datasets with class noise. Third, as mentioned in Section \ref{RobustoCND}, RD-GBG considers noise elimination, so it still performs exceptionally well in scenarios with class imbalance and noise.

\subsection{Parameter Sensitivity Analysis}

This section primarily validates the impact of different values of the density tolerance $\rho$ in the GBABS, including the sampling ratio and the quality of the sampled dataset. The quality of the sampled dataset is verified through the performance of a classifier, without loss of generality, where the classifier used is DT.

\begin{figure}[htbp]
\centering
\includegraphics[height=2.0in,width=3.5in]{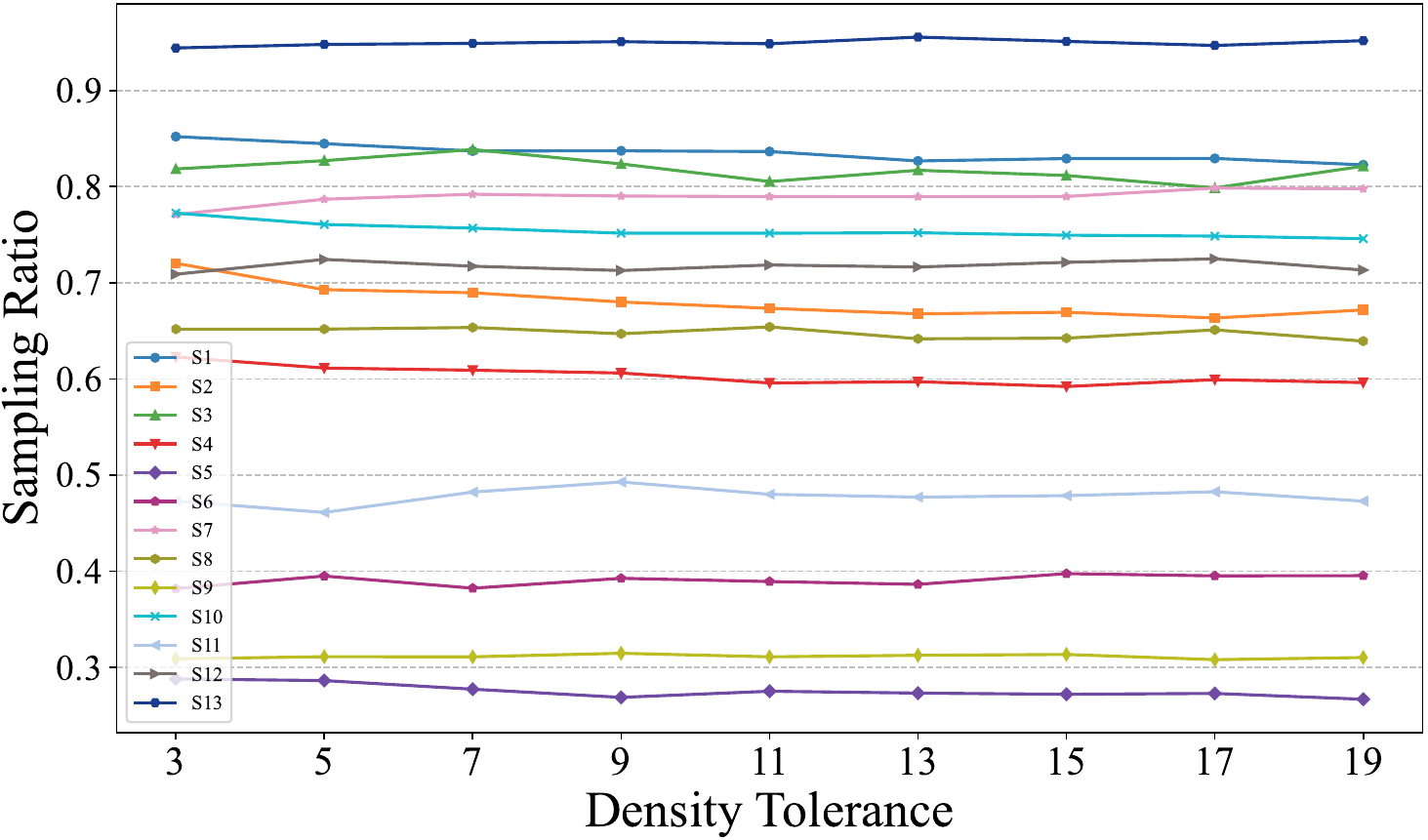}
\caption {Impact of density tolerance $\rho$ on sampling ratio.}
\label{fig8}
\end{figure}

\begin{figure}[htbp]
\centering
\includegraphics[height=2.0in,width=3.5in]{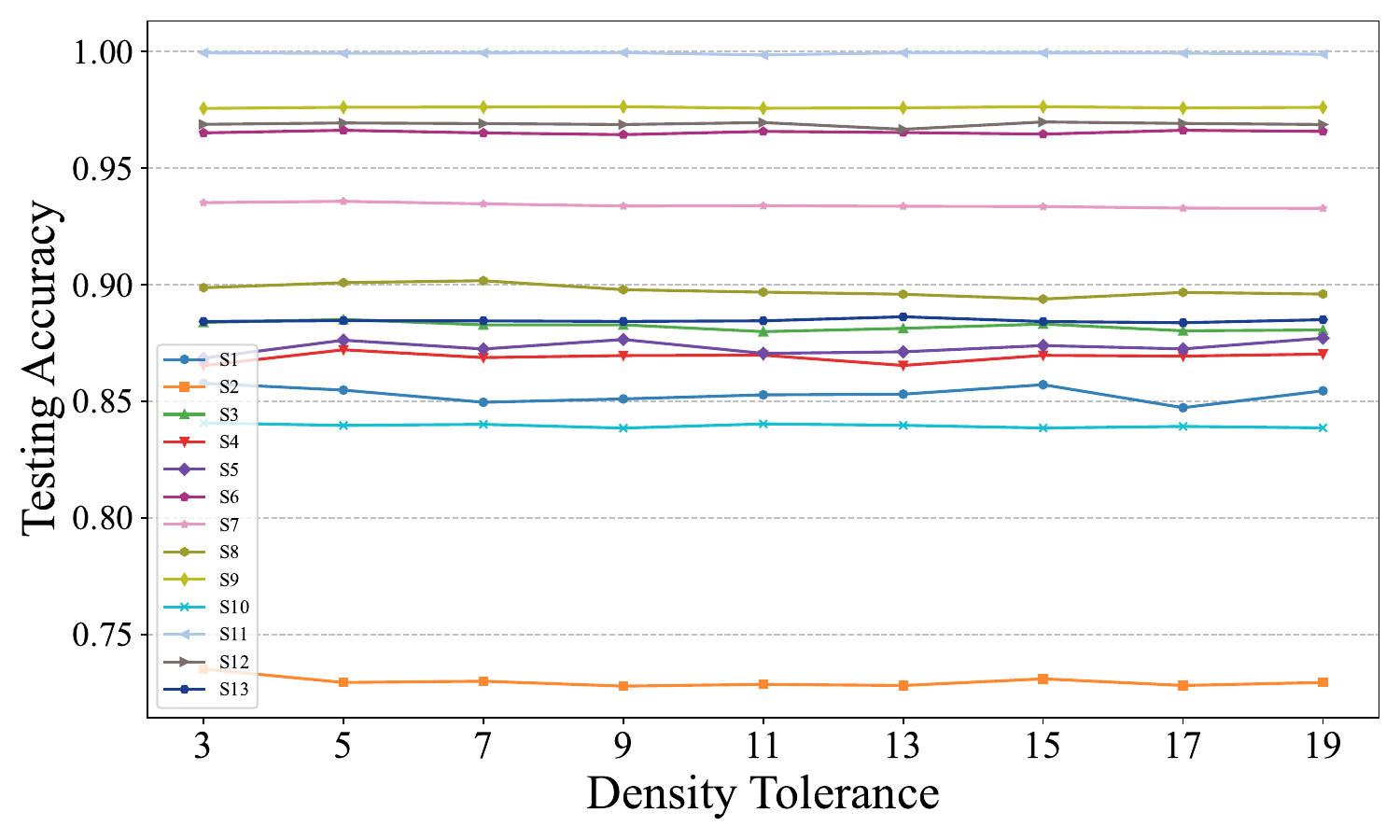}
\caption {Impact of density tolerance $\rho$ on testing $Accuracy$ of DT.}
\label{fig9}
\end{figure}

Fig.\ref{fig8} illustrates the sampling ratios of GBABS for all standard datasets listed in Table\ref{table1}, when the density tolerance $\rho$ takes values of $3, 5, 7, 9, 11, 13, 15, 17$, and $19$. Fig.\ref{fig9} shows the corresponding testing $Accuracy$ of GBABS-based DT.  According to Fig.\ref{fig8}, as the value of $\rho$ increases, the sampling ratio of GBABS tends to stabilize across all datasets. Meanwhile, as shown in Fig.\ref{fig9}, the testing $Accuracy$ of GBABS-based DT shows no significant variation with $\rho$, especially for datasets with larger sample sizes and higher dimensions. As a result, GBABS exhibits insensitivity to its hyperparameter.

\section{Conclusion}
\label{sec: conclusion}
This paper proposes an approximate borderline sampling method using GBs, incorporating RD-GBG and GBABS, which extends borderline sampling to a more general setting with a linear time complexity that accelerates classifiers. Notably, the RD-GBG method addresses a major limitation of existing GB-based sampling approaches by eliminating GB overlap and redefining GBs, ensuring a closer alignment between the generated GBs and the original dataset distribution. The GBABS method further advances the sampling process by adaptively identifying borderline samples, effectively reducing redundancy and noise while mitigating class imbalance effects. Experimental results confirm that GBABS reduces the sampling ratio while maintaining sample quality, improving the efficiency, robustness, and performance of classifiers. However, the time complexity of the GBABS is not ideal when facing high-dimensional feature spaces. Future work will focus on improving its efficiency to enable broader applicability.

\ifCLASSOPTIONcaptionsoff
  \newpage
\fi

\bibliography{mybibfile}

\end{document}